%% file: main.tex
\pdfoutput=1

\documentclass[11pt]{article}

\usepackage{acl}

\usepackage{times}
\usepackage{latexsym}

\usepackage[T1]{fontenc}
\usepackage{graphicx}
\usepackage[utf8]{inputenc}

\usepackage{microtype}
\usepackage{url}
\usepackage{amsmath}
\usepackage{amssymb}
\usepackage[noend]{algpseudocode}
\usepackage{algorithm}
\usepackage{xcolor}
\usepackage{booktabs}
\usepackage{adjustbox}
\usepackage[font=small,labelfont=bf]{caption}
\usepackage{subcaption}
\usepackage{CJKutf8}
\usepackage[normalem]{ulem}
\usepackage{xcolor,colortbl}
\usepackage{arydshln}

\usepackage{empheq}
\usepackage[most]{tcolorbox}

\newtcbox{\mymath}[1][]{%
    nobeforeafter, math upper, tcbox raise base,
    enhanced, colframe=blue!30!black,
    colback=blue!30, boxrule=1pt,
    #1}
\input{utils/macro.tex}

\title{Hierarchical Phrase-based Sequence-to-Sequence Learning}

\author{
Bailin Wang\\
MIT \\
\texttt{\fontsize{10pt}{10pt}\selectfont bailinw@mit.edu}\\ \And
Ivan Titov\\
University of Edinburgh\\University of Amsterdam\\
\texttt{\fontsize{10pt}{10pt}\selectfont ititov@inf.ed.ac.uk} \\ \And
Jacob Andreas\\
MIT \\
\texttt{\fontsize{10pt}{10pt}\selectfont jda@mit.edu} \\   \And
Yoon Kim\\
MIT \\
\texttt{\fontsize{10pt}{10pt}\selectfont yoonkim@mit.edu} \\
}

\begin{document}
\setlength{\abovedisplayskip}{4pt}
\setlength{\belowdisplayskip}{4pt}
\setlength{\abovedisplayshortskip}{4pt}
\setlength{\belowdisplayshortskip}{4pt}

\maketitle

\begin{abstract}
\vspace{-1.5mm}
We describe a neural transducer that maintains the flexibility  of standard sequence-to-sequence (seq2seq) models while incorporating hierarchical phrases as a source of inductive bias during training and as explicit constraints during inference. Our approach trains two models: a discriminative parser based on a  bracketing transduction grammar whose derivation tree hierarchically aligns source and target phrases, and a neural seq2seq model that learns to translate the aligned phrases one-by-one. 
We use the same seq2seq model to translate at all phrase scales, which results in two  inference modes: one mode in which the parser is discarded and only the seq2seq component is used at the sequence-level, and another in which the parser is combined with the seq2seq model. Decoding in the latter mode is done with the cube-pruned CKY algorithm, which is more involved but can make use of  new translation rules during inference. We formalize our model as a source-conditioned synchronous grammar and develop an efficient variational inference algorithm for training.
When applied on top of both randomly initialized and pretrained seq2seq models, we find that both inference modes performs well compared to baselines on small scale machine translation benchmarks.
 \blfootnote{\noindent \hspace{-6mm} Code available at: \url{https://github.com/berlino/btg-seq2seq}.}
\vspace{-2.5mm}
\end{abstract}
\vspace{-2mm}
\section{Introduction}
\vspace{-2mm}
\begin{figure}[t]
        \vspace{-1mm}
    \centering
    \includegraphics[scale=0.20]{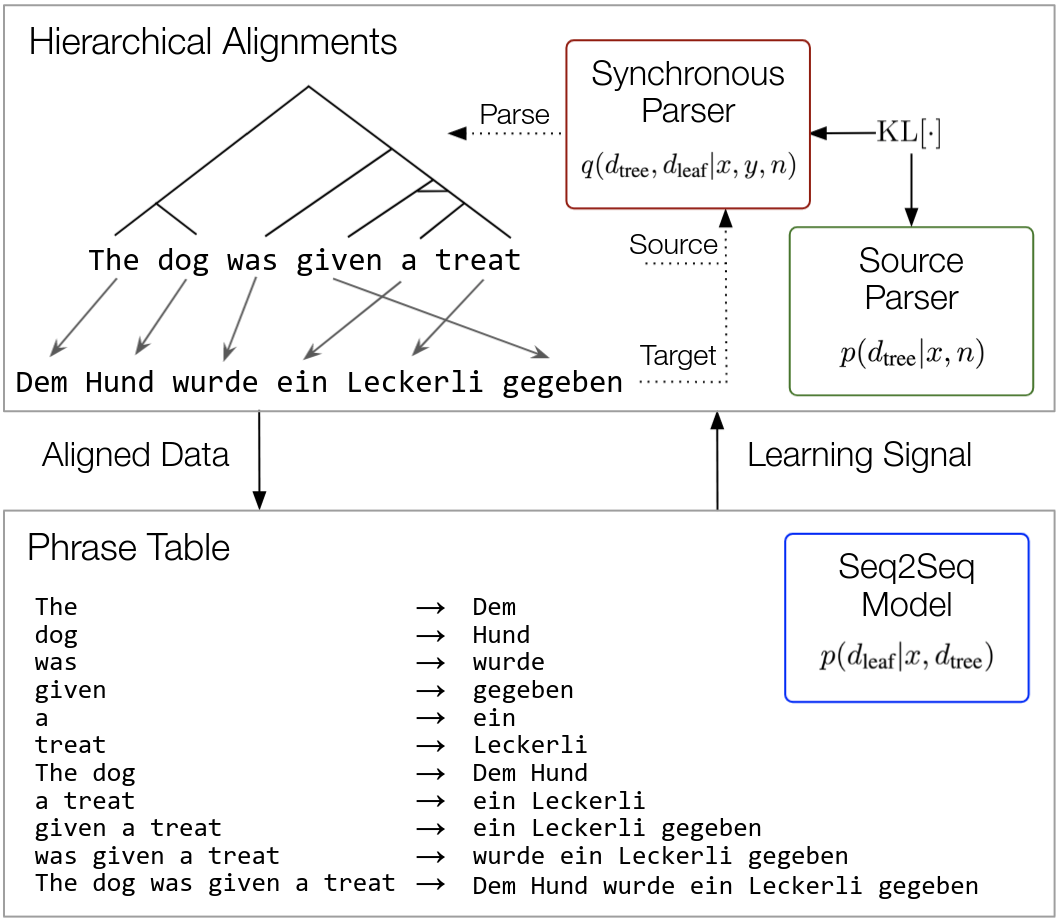}
        \vspace{-3mm}
    \caption{An overview of our approach. During training a variational synchronous parser hierarchically segments and aligns source-target phrases (top), which provides training pairs for a seq2seq model (bottom).  The seq2seq loss provides learning signal to the  synchronous parser which subsequently provides signal to  (and is regularized by) a monolingual source parser. After training, there are two inference models: one mode in which both parsers are discarded, which results in a regular seq2seq model that has been regularized by latent hierarchical phrase pairs; another mode in which the seq2seq model can be combined with the source parser to derive a neural synchronous grammar. Inference in this mode is done via translation-as-parsing with cube-pruned CKY.}
    \label{fig:overview}
    \vspace{-6mm}
\end{figure}

Despite the improvements in performance and data-efficiency enabled by recent advances in pretraining, 
standard neural sequence-to-sequence (seq2seq) models can still fail to model the hierarchical structure of sentences 
and be  brittle with respect to to novel syntactic structures \cite{lake2018generalization,kim-linzen-2020-cogs,weibenhorn2022}. 
It has moreover not been clear how to incorporate explicit constraints such as translation rules (e.g., for translating idioms) 
into these black-box models without changing the underlying architecture.  Classic grammar- and automaton-based approaches
are well-suited for capturing hierarchical structure and can readily incorporate new rules, 
but have trouble with examples that do not conform exactly to the  symbolic specifications. 
This paper describes a neural transducer that maintains the flexibility of neural seq2seq models  
but  still models hierarchical phrase alignments between source and target sentences, 
which have been shown to be useful for transduction tasks \cite[\textit{inter alia}]{chiang-2005-hierarchical,wong-mooney-2007-learning}.  
This transducer bottoms out in an ordinary neural seq2seq model,  
and can thus take advantage of pre-training schemes like BART \cite{lewis-etal-2020-bart,liu-etal-2020-multilingual-denoising} and T5 \cite{raffel2020t5,xue-etal-2021-mt5}.

The transducer consists of two components learned end-to-end: a discriminative parser based on a  bracketing transduction grammar~\cite[BTG;][]{wu-1997-stochastic} 
whose derivation tree hierarchically aligns source and target phrases, and a neural seq2seq model that learns to translate the aligned phrases one-by-one.  
Importantly, a single seq2seq model is trained to translate phrases at \textit{all scales}, including at the sentence-level, which results in two inference modes. 
In the first mode, we discard the parser and simply decode with the resulting seq2seq model at the sequence level, which maintains fast inference but 
still makes use of a seq2seq model that has been exposed to (and regularized by) latent hierarchical phrase alignments. 
This approach can be seen a form of \textit{learned} data augmentation \cite{akyurek2021learning}, wherein a model is guided away from wrong generalizations 
by  additionally being trained on crafted data that coarsely express the desired inductive biases. 
In the second mode, we view the combined parser and seq2seq model as a source-conditioned neural synchronous grammar 
to derive a set of synchronous grammar rules along with their scores (a ``neural phrase table''), 
and decode with a modified version of the classic cube-pruned CKY algorithm \cite{huang-chiang-2005-better}.  
While more involved, this decoding scheme can incorporate explicit constraints and new translation rules during inference.

We formulate our approach as a latent variable model and use variational learning to efficiently maximize a lower bound on the log marginal likelihood. 
This results in an intuitive training scheme wherein the seq2seq component is trained on phrase tables derived from hierarchical alignments sampled from a variational posterior synchronous parser, as shown in Figure~\ref{fig:overview}. 
When applied on top of both randomly initialized and pretrained seq2seq models across various small-scale machine translation benchmarks, we find that both modes improve upon baseline seq2seq models.

\input{sections/method.tex}

\input{sections/experiment.tex}

\input{sections/related_work.tex}

\input{sections/conclusion}

\input{sections/limitations.tex}

\vspace{-4mm}
\bibliography{bib/anthology,bib/main}
\bibliographystyle{acl_natbib}
\appendix
\newpage

\input{sections/appendix.tex}
\end{document}

%% file: utils/macro.tex
\newcommand{\src}{x}
\newcommand{\tgt}{y}
\newcommand{\dtree}{d_{\text{tree}}}
\newcommand{\dleaf}{d_{\text{leaf}}}

\newcommand{\given}{\,\vert\,}

\newcommand{\mcR}{\mathcal{R}}

\newcommand{\E}{\mathbb{E}}
\newcommand{\ENT}{\mathbb{H}}

\newcommand{\chart}{\mathbb{C}}

\newcommand\blfootnote[1]{%
  \begingroup
  \renewcommand\thefootnote{}\footnote{#1}%
  \addtocounter{footnote}{-1}%
  \endgroup
}

\DeclareMathOperator*{\KL}{KL}

\DeclareMathOperator*{\yield}{yield}

\DeclareMathOperator*{\argmax}{argmax}

\DeclareMathOperator*{\Enc}{Encoder}
\DeclareMathOperator*{\score}{\pi}
\DeclareMathOperator*{\pscore}{\tilde\pi}

\algnewcommand{\LineComment}[1]{\State \(\triangleright\) #1}

\definecolor{Gray}{gray}{0.9}

%% file: sections/method.tex
\vspace{-2mm}
\section{Approach}
\vspace{-2.5mm}
\begin{figure*}
    \vspace{-2.5mm}
    \centering
    \includegraphics[scale=0.175]{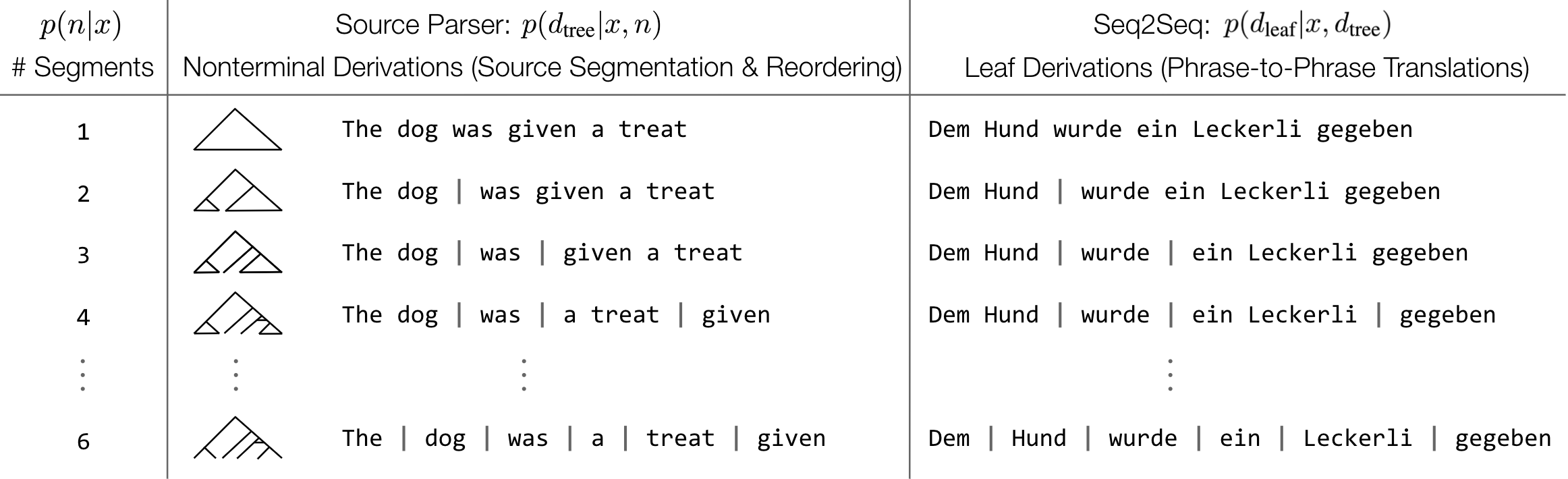}
    \vspace{-3mm}
    \caption{Our seq2seq-based synchronous grammar defines a distribution over target derivations $d$ given source string $\src$ by decomposing $p(d | \src)$ into three components,  $p(d | \src) = p(n|x)p(\dtree | x, n)p(\dleaf| x,\dtree)$. 
    We first sample the number of segments in the target $n \sim p(n|x)$.  Then we sample the tree topology $\dtree \sim p(\dtree | x, n)$, which gives a segmentation and reordering of the source sequence. 
    Finally, the leaf derivations are given by translating the segmented and reordered source phrases one-by-one with a seq2seq model $p(\dleaf| x,\dtree)$, where the phrase-by-phrase translations are  contextualized against the entire source sentence. We show some example trees $\dtree$ and leaf derivations $\dleaf$ for various values of $n$. Note that each $n$ could have multiple segmentations and reorderings.}
    \label{fig:grammar}
        \vspace{-5.5mm}
\end{figure*}

\paragraph{Notation.} 
We use $\src, \tgt$ to denote the source/target strings, 
$\src_{i:k}, \tgt_{a:c}$ to denote the source/target spans, and $x_i$ to denote the token with index $i$.
\vspace{-1mm}
\subsection{A Seq2Seq-Based Synchronous Grammar}
\vspace{-1mm}
The inversion transduction grammar (ITG), introduced by \citet{wu-1997-stochastic}, defines a synchronous grammar that hierarchically generates source and target strings in tandem. 
The bracketing transduction grammar (BTG) is a simplified version of ITG with only one nonterminal, with the primary goal of modeling alignments between source and target words. 
We propose to use a variant of BTG which defines a stochastic monolingual grammar
over target strings conditioned on  source strings and can use a neural seq2seq model to generate phrases.  

As a motivating example, consider the following English-Chinese example from~\citet{shao-etal-2018-evaluating}:\footnote{The phrase-level translations are:  
  \begin{CJK*}{UTF8}{gbsn} 他们  \end{CJK*} $\rightarrow$ \textit{They}, 
  \begin{CJK*}{UTF8}{gbsn} 在 $\,\,$ 桥上  \end{CJK*} $\rightarrow$ \textit{were on the bridge}, 
\begin{CJK*}{UTF8}{gbsn} 谈笑 $\,\,$ 风生   \end{CJK*} $\rightarrow$ \textit{laughing it up}.}
\begin{align*}
\vspace{-2mm}
   &\begin{CJK*}{UTF8}{gbsn}\text{他们 $\,\,$ 在 $\,\,$ 桥上 $\,\,$ 谈笑 $\,\,$ 风生}\end{CJK*} \rightarrow 
  \\ &\textit{They were laughing it up on the bridge}
  \vspace{-2mm}
\end{align*}
A translator for this sentence is faced with several problems: first, they must identify the appropriate unit of translation (number of phrases); two, they must reorder phrases if necessary; finally, they must be able translate phrase-by-phrase, taking into account phenomena such as idioms whose translations cannot be obtained by stitching together translations of subparts (e.g.,   \begin{CJK*}{UTF8}{gbsn} 谈笑 $\,\,$ 风生   \end{CJK*} $\rightarrow$ \textit{laughing it up}), while also making sure the final translation is fluent as a whole.
We model this process with a \textit{source-conditioned} synchronous grammar whose target derivation probabilistically reifies the above process. Informally, this grammar first samples the number of segments to generate on the target ($n$),  then segments and reorders source phrases ($\dtree$), and finally translates phrase-by-phrase while still conditioning on the full source sentence for  fluent translations ($\dleaf$). This is shown in Figure~\ref{fig:grammar}.

More formally, this  monolingual grammar over the target can be represented by a tuple
\begin{align*}
    G[\src] = \left(Root, \{T^n, S^n, I^n\}_{n=1}^{|x|}, C, \Sigma, \mcR[\src]\right),
\end{align*} 
where $Root$ is the distinguished start symbol, $\{T^n, S^n, I^n\}_{n=1}^{|x|}$  are nonterminals where $n$ denotes the number of segments associated with the nonterminal (to be explained below), $|x|$ is the source sentence length, $C$ is a special nonterminal that emits phrases, and $\Sigma$ is the target alphabet. 
The rule set $\mcR[\src]$ is given by a fixed set of context-free production rules. For the root node we have $\forall n \in \{1, 2 \dots |x|\}$,
\begin{align*}
Root \to T^n,  \hspace{4mm}  T^n \to S^n[x], \hspace{4mm} T^n \to I^n[x], 
\end{align*}
where $S, I$ represent the $S$traight and $I$nverted nonterminals. Binary nonterminal rules are given by,
\begin{align*}
    & \forall i,j,k \in \{0, 1 \dots |x|\} && \textrm{s.t.} \ \  i < j < k \\
        & \hspace{3mm} S^n[x_{i:k}] \to I^l[x_{i:j}] S^r[x_{j:k}], \\ 
        & \hspace{3mm} S^n[x_{i:k}] \to I^l[x_{i:j}] I^r[x_{j:k}],   \\
    & \hspace{3mm} I^n[x_{i:k}] \to S^l[x_{j:k}] I^r[x_{i:j}],    \\ 
    & \hspace{3mm} I^n[x_{i:k}] \to S^l[x_{j:k}] S^r[x_{i:j}],  
    \end{align*}
where each nonterminal is indexed by (i.e., aligned to) a source span $x_{i:k}$, and the superscript $n$ refers to the number of segments to generate.\footnote{
Rules  of the form $S \to S I$ and $I \to I S$ are disallowed. This resolves ambiguities where certain reorderings can be encoded by multiple derivations.} 
We force $l+r=n$ to ensure that the number of segments generated by the $l$eft and $r$ight nonterminal is equal to $n$. Finally, if $S^1$ or $I^1$ is generated (i.e., there is only one segment to be generated), we move to the leaf nonterminal $C$  and generate target phrases,
    \begin{align*}
    & \hspace{3mm} S^1[x_{i:k}] \to C[x_{i:k}], && I^1[x_{i:k}] \to C[x_{i:k}], \\
    & \hspace{3mm} C[x_{i:k}] \to v, && v \in \Sigma^{+}.  
\end{align*}
We use a probabilistic version of the above grammar where each rule is weighted via $\score: \mcR \to \mathbb R_{\geq 0} $. Importantly, for the leaf nonterminal we use a (potentially pretrained) neural seq2seq model, i.e.,
\begin{align*}
    \score(C[x_{i:k}] \to v) = p_{\text{seq2seq}}(v | x_{i:k}). 
\end{align*}
Foreshadowing, if we force $Root \to T^1$ in the beginning of derivation then this  corresponds to the usual seq2seq approach which generates the full target sequence conditioned on the source sequence.

\vspace{-1mm}
\subsection{Parameterization}
\vspace{-1mm}
By assigning scores to rules conditioned on $\src$ with function $\score$, we can obtain a distribution $p(d|\src)$ over 
all possible derivations $d$ licensed by $G[\src]$.
Note that $d$ is fully characterized by: (1) $n$, the total number of target segments, (2) $\dtree$, the set of derivation rules excluding the leaf rules (i.e., rules with $C[x_{i:k}]$ on the left hand side),
and (3) $\dleaf$, the set of leaf derivation rules. 
We hence decompose the derivation into three distributions, 
$p(d|\src) = p(n|\src) p(\dtree|\src, n)p(\dleaf| \src, \dtree)$. 
Intuitively, $\dtree$ encodes the segmentations and reorderings of $x$, 
while $\dleaf$ encodes the phrase translations which implicitly segment $y$.

\vspace{-1mm}
\paragraph{$\boldsymbol{p(n|\src)}$.} 
For the distribution over the number of target segments we use a geometric distribution,
\begin{equation*}
p(n|\src)=\lambda(1 - \lambda)^{n-1},
\end{equation*}
where $0 < \lambda < 1$, $1 \leq n < |\src|$. 
This sets the probability of the upper bound $n = |\src|$ to be $(1 - \lambda)^{|\src| - 1}$. 
This distribution is implemented by the following configuration of the grammar:
\begin{align*}
& \score(Root \to T^n) = p(n|\src).
\end{align*}
The geometric distribution favors fewer target segments, which aligns with one of our goals which is  
to be able to use the resulting model as a regular seq2seq system (i.e., forcing $Root \to T^1$).

\vspace{-1mm}
\paragraph{$\boldsymbol{{p(\dtree|\src, n)}}$.}
For the distribution over source segmentations and reorderings, 
we use a discriminative CRF parser whose scores are given by neural features over spans. 
The score for rule $ R \in \mcR$, e.g., $S^n[x_{i:k}] \to I^l[x_{i:j}] A^r[x_{j:k}]$, is given by, 
\begin{align*}
\score(R) = \exp \left( e_{A}^\top f(\mathbf{\tilde h}_{i:j}, \mathbf{\tilde h}_{j:k}) \right), 
\end{align*}
where $A \in \{S, I\}$, $\mathbf{\tilde h}_{i:j}$ and $\mathbf{\tilde h}_{j:k}$ are the span features  derived from the contextualized word representations of tokens at $i, j, k$ from a Transformer encoder as in \citep{kitaev2018parsing}, 
$f$ is an MLP that takes into two span features, 
and $e_{A}$ is a one-hot vector that extracts the corresponding scalar score. (Hence we use the same score  all $n$).  These scores are globally normalized, i.e.,
\begin{align*}
    p(\dtree | x, n) \propto \prod_{R \in \dtree} \score(R),
\end{align*}
to obtain a distribution over $\dtree$, where $R $ are the rules in $\dtree$. Algorithms~\ref{algo:inside-tree}, \ref{algo:sample-tree} in appendix~\ref{app:algo} give the dynamic programs for normalization/sampling.

\paragraph{$\boldsymbol{p(\dleaf| \src, \dtree)}$.} Finally, for  the distribution over the leaf translations, the phrase transduction rule probabilities $p(C[x_{i:k}] \to v)$ are parameterized 
with a (potentially pretrained) neural seq2seq model. For the seq2seq component, 
instead of just conditioning on $x_{i:k}$ to parameterize $p_{\text{seq2seq}}(v | x_{i:k})$, 
in practice we use the contextualized representation of $x_{i:k}$ from a Transformer encoder 
that conditions on the full source sentence. Hence, our approach can be seen learning \textit{contextualized} phrase-to-phrase translation rules. Specifically, for a length-$m$ target phrase we have,
\begin{align*}
    & p(\dleaf| \src, \dtree) = \prod_{R \in \dleaf} \score(R) \\
    &\score(C[x_{i:k}] \to v) = \prod_{t = 1}^m p_{\text{seq2seq}} (v_t \given v_{<t}, \mathbf{h}_{i:k}), \\
    &\mathbf{h}_{i:k} = \Enc(x)_{i:k}, 
\end{align*}
where $\mathbf{h}_{i:k}$ refers to the contextualized representation of $x_{i:k}$ that is obtained after passing it through a Transformer encoder. The decoder attends over $\mathbf{h}_{i:k}$ and $v_{<t}$ to produce a distribution over $v_t$. The two ``exit'' rules 
are assigned  a  score $\score(A^1[x_{i:k}] \to C[x_{i:k}]) = 1$ where $A \in \{S, I\}$.
\vspace{-1mm}
\paragraph{Remark.}
This grammar is a mixture of both locally and globally normalized models. In particular, $p(n|\src)$ and $p(\dleaf|\src, \dtree)$
are locally normalized as the scores for these rules are already  probabilities.   In contrast,  $p(\dtree|\src, n)$ is obtained by globally normalizing the score of $\dtree$ with respect 
to a normalization constant $Z_{\text{tree}}(\src, n)$.\footnote{Locally normalized parameterization of $\dtree$ would require $\score$
to take into account the number of segments $n$ (and $l, r$), which would lead to a more complex rule-scoring function. 
In our parameterization $\score(R)$ does not depend on the number of segments, and $n, l, r$ only affect
$p(\dtree|\src, n)$ via the normalization $Z_{\text{tree}}(\src, n)$.} In the  appendix we show how to convert the unnormalized
scores $\score$ in the globally normalized model into normalized probabilities $\pscore$ (Algorithm~\ref{algo:inside-tree}), which will be used for CKY decoding in Section~\ref{subsec:infer}.

We also share the parameters between components whenever possible. For example, the encoder of the seq2seq component is the same encoder that is used to derive span features for $\dtree$. The decoder of the seq2seq component is used as part of the variational distribution when deriving span features for the variational parser. As we will primarily be working with small datasets, such parameter sharing will provide implicit regularization to each of the components of our grammar and militate against overfitting. We provide the full parameterization of each component in  appendix~\ref{app:parameterization}.

\vspace{-2mm}
\subsection{Learning}
\vspace{-1mm}
Our model defines a distribution over target trees $d = (n, \dtree, \dleaf)$ (and by marginalization, target strings $y$) conditioned on a source string $\src$, \vspace{-1mm}
\begin{align*}
      &  p(\tgt|\src) =  \sum_{d \in D(\src, \tgt)} p(d|\src) \\
      &= \sum_{d \in D(\src, \tgt)} p(n|\src) p(\dtree | \src, n) p(\dleaf | \src, \dtree), 
\vspace{-3cm}
\end{align*}
where $D(x,y)$ is the set of trees such that $\yield(D(x, y)) = (x,y)$. While this marginalization can  theoretically be done in $O(L^7)$ time 
where $L = \min(|x|, |y|)$,\footnote{For each  $n \in [L]$, we must run the $O(n^6)$ bitext inside algorithm, so the total runtime is $\sum_{n=1}^L O(n^6) = O(L^7)$.} we found it impractical in practice. We instead  to use variational approximations to the posteriors of $n$, $\dtree$, $\dleaf$ and optimize
a lower bound on the log marginal likelihood whose gradient estimator can be obtained in $O(L^3)$ time. 
Each of the three variational distributions $q$  is analogous to the three distributions $p$  introduced previously, but additionally conditions on  $\tgt$.
\vspace{-1mm}
\paragraph{$\boldsymbol{q(n | \src, \tgt)}$.}
Similar to $p(n|x)$, we use a geometric distribution to model $q(n|\src, \tgt)$. The only difference 
is that we have $1 \leq n \leq \min(|\src|, |\tgt|)$.
\vspace{-3mm}
\paragraph{$\boldsymbol{q(\dtree | \src, \tgt, n)}$.}
This is parameterized with neural span scores  over $\src$ in the same way as in $p(\dtree | \src, n)$, 
except that the span features also condition on the target $\tgt$. 
That is,  $\mathbf{\tilde h}_{i:j}$,  $\mathbf{\tilde h}_{j:k}$ are now the difference features of the \textit{decoder}'s contextualized representations 
over $\src$ from a Transformer encoder-decoder, where $\tgt$ is given to the encoder.\footnote{We use this parameterization (instead of, e.g., a bidirectional Transformer encoder over the concatenation of $\src$ and $\tgt$) 
in order to take advantage of pretrained multilingual encoder-decoder models such as mBART.} The dynamic programs for normalization/sampling in this CRF is the same as in ${p(\dtree|\src, n)}$. 
\vspace{-1mm}
\paragraph{$\boldsymbol{q(\dleaf | x, y, \dtree)}$.}
Observe that conditioned on $(x, y, \dtree)$, the only source of uncertainty in $\dleaf$ is $C[x_{i:k}] \to y_{a:c}$ for spans $y_{a:c}$. That is, 
$q(\dleaf | x, y, \dtree)$ is effectively a hierarchical segmentation model with a fixed topology given by $\dtree$. 
The segmentation model can be described by a probabilistic parser over the following grammar,
\begin{align*}
& G[\tgt, \dtree] = \left(Root, \{T^s\}_{s=1}^{n}, \mcR[\tgt, \dtree]\right),
\end{align*}
where rule set $\mcR[\tgt, \dtree]$ includes (here $m = l+r)$,
\begin{align*}
& Root \to T^n[y], \ 
T^m[y_{a:c}] \to T^l[y_{a:b}] T^r[y_{b:c}]
\end{align*}
where the second rule exists only when the rule like $A^m \to B^l C^r$ is in $\dtree$.
We use span features over $\tgt$ for parameterization similarly to  $p(\dtree | \src, n)$.
Algorithms~\ref{algo:inside-leaf},~\ref{algo:sample-leaf} in appendix~\ref{app:algo} give the normalization/sampling dynamic programs for this CRF.
\vspace{-2mm}
\paragraph{Optimization.} With the variational distributions in hand, we are now ready to give the  following  evidence lower bound (ignoring constant terms),
\vspace{-1mm}
\begin{align*}
\vspace{-2mm}
   & \log p(\tgt | \src) \geq \E_{ q(n | \src, \tgt) }\Big \{  
         \E_{q(\dtree | \src, \tgt, n)}  \Big[ \\
         & \,\,\,\,\E_{q(\dleaf | x, y, \dtree)}[\log  p(\dleaf | \src, \dtree)]\\& \,\,\,\, + \ENT[q(\dleaf | x, y, \dtree)]\Big]   \\
    & \,\, - \KL [q(\dtree | \src, \tgt, n) \, \Vert \, p(\dtree | \src, n) ] \Big \}.
\end{align*}
(The $\KL$ between $q(n|x,y)$ and $p(n|x)$ is a constant and hence omitted.) See appendix~\ref{app:elbo} for the derivation.
While the above objective  seems involved, in practice it results in an intuitive training scheme. Let $\theta$ be the grammar parameters and $\phi$ be the variational parameters.\footnote{We share the Transformer encoder/decoder between $p_\theta(\cdot)$ and $q_\phi(\cdot)$, and therefore the only parameters that are different between $\theta$ and $\phi$ are the MLPs' parameters for scoring rules in the CRF parsers.} Training proceeds as follows:
\begin{enumerate}
\item Sample number of target segments, 
\begin{align*}n \sim q(n | \src, \tgt). \end{align*}
\item Sample trees from variational parsers,
\begin{align*} &d'_{\text{tree}} \sim q(\dtree | \src, \tgt, n), \\
&d'_{\text{leaf}} \sim q(\dleaf | \src, \tgt,d'_{\text{tree}}),
\end{align*}
and obtain the set of hierarchical phrase alignments (i.e., phrase table) $A$ implied by $(d'_{\text{tree}}, d'_{\text{leaf}})$. 
\item Train seq2seq model by backpropagating the leaf likelihood $\log p(d'_{\text{leaf}} | x, d'_{\text{tree}})$ to the seq2seq model,
\begin{align*} 
    \sum_{(x_{i:k}, y_{a:c}) \in A} \nabla_\theta \log p_{\text{seq2seq}}(y_{a:c} | x_{i:k}).
\end{align*}
\item {Train variational parsers} by backpropagating the score function and KL objectives with respect to $q(\dtree | x, y, n)$,
\begin{align*}
&\hspace{-2mm} \left(\log p(d'_{\text{leaf}} | x, d'_{\text{tree}}) + \ENT[q(d_{\dleaf} | x, y, d'_{\text{tree}})] \right)   \\ & \,\,\times \nabla_\phi \log q(d'_{\text{tree}}|x,y, n) - \nabla_\phi\KL[\cdot],
\end{align*}
and also with respect to $q(\dleaf | x, y, n)$,\vspace{2mm}
\begin{align*}
&\hspace{-2mm}  \log p(d'_{\text{leaf}} | x, d'_{\text{tree}})\times \nabla_\phi \log q(d'_{\text{leaf}}|x,y, d'_{\text{tree}}) \\ &\,\, + \nabla_\phi\ENT[q(\dleaf | x, y, d'_{\text{tree}})].
\end{align*}
\item {Train source  parser} by backpropagating $- \nabla_\theta \KL[\cdot]$  to $p(\dtree | x, n)$.
\end{enumerate}
For the score function gradient estimators we also employ a self-critical control-variate \citep{rennie2017self} with argmax trees.The KL and entropy metrics are calculated exactly with the inside algorithm with  the appropriate semirings \cite{li:2009}. This training scheme is shown in Figure~\ref{fig:overview}. See Algorithms~\ref{algo:leaf-entropy},~\ref{algo:tree-kl} in appendix~\ref{app:algo} for the entropy/KL dynamic programs.

\vspace{-1mm}
\paragraph{Complexity.}  
The above training scheme reduces time complexity from $O(L^7)$ to $O(L^3)$ since we 
 do not marginalize over all possible number of segments and instead sample $n$, and further decompose the derivation tree into 
$(\dtree, \dleaf)$ and sample from CRF parsers, which takes $O(L^3)$. The KL/entropy calculations are also $O(L^3)$.

\vspace{-2mm}
\subsection{Inference: Two Decoding Modes}
\label{subsec:infer}
\vspace{-2mm}
During inference, we can use the seq2seq to either directly decode 
at the sentence-level (i.e.,  setting $n=1$ and choosing $Root \to T^1$), or generate phrase-level translations and compose them together. 
The first decoding strategy maintains the efficiency of seq2seq, and variational learning can be seen as a structured training algorithm that regularizes the overly flexible seq2seq model with latent phrase alignments.
The second decoding strategy takes into account translations at all scales and can further incorporate constraints, such as new translation rules during decoding.

\vspace{-1mm}
\paragraph*{Seq2Seq decoding.}
Setting $n=1$, decoding in $G[\src]$ this case is reduced to the 
standard  sentence-level seq2seq decoding for the terminal 
rules $C[x_{0, |x|}] \to v$ with beam search. We call this variant of the model  \textbf{BTG-$1$ Seq2Seq}.

\paragraph*{CKY decoding.} Here we aim to find $y$ with maximal marginal probability $\argmax_y \sum_{d \in T(x, y)} p(d|x)$.
This requires exploring all possible derivations licensed by $G[\src]$.
We  employ CKY decoding algorithm based on the following recursion:
\vspace{-1mm}
\begin{align*}
\vspace{-2mm}
    & \chart_{i:k}[v_1v_2, S^n]  = \sum_{l = 1, r = n-l}^{n-1} \big \{ \\
        & \hspace{7mm} \chart_{i:j}[v_1, I^l] \cdot \chart_{j:k}[v_2, S^r] \cdot \pscore({\scriptstyle S^n \to I^l S^r}) + \\
        & \hspace{6.5mm} \chart_{i:j}[v_1, I^l] \cdot \chart_{j:k}[v_2, I^r] \cdot \pscore({\scriptstyle S^n \to I^l I^r}) \, \big \} \\
    \vspace{2mm}
    & \chart_{i:k}[v_1v_2, I^n]  = \sum_{l = 1, r = n-l}^{n-1} \big \{ \\
        & \hspace{7mm} \chart_{j:k}[v_1, S^l] \cdot \chart_{i:j}[v_2, I^r] \cdot \pscore({\scriptstyle I^n \to S^l I^r}) + \\
        & \hspace{6.5mm} \chart_{j:k}[v_1, S^l] \cdot \chart_{i:j}[v_2, S^r]  \cdot \pscore({\scriptstyle I^n \to S^l S^r}) \, \big \} 
\vspace{-10mm}
\end{align*}
where $\chart_{i:k}[v, A]$ stores the score of generating target phrase $v$ based on source phrase $\src_{i:k}$ with nonterminal $A$. In the above we abbreviate $\pscore({\scriptstyle S^n \to I^l S^r})$ to mean $\pscore({\scriptstyle S^n[x_{i:k}] \to I^l[x_{i:j}] S^r[x_{j:k}]})$ since $i,j,k$ are implied by the $\chart$ terms that this quantity is multiplied with.
Here $v, v_1, v_2$ denotes target phrases, and $v_1 v_2$ refers to the concatenation of two phrases.
The chart is initialized by phrase pairs that are generated based on seq2seq,
\begin{align*}
    & \chart_{i:k}[v, S^1]  = p_{\text{seq2seq}}(v|x_{i:k}) \cdot \score({\scriptstyle S^1 \to C}), \\
    & \chart_{i:k}[v, I^1]  = p_{\text{seq2seq}}(v|x_{i:k}) \cdot \score({\scriptstyle I^1 \to C})
\end{align*}
The final prediction is based on the chart item for the root node computed via:
\begin{align*}
\vspace{-2mm}
    & \chart[v, T^n]  = \chart_{0, |\src|}[v, S^n] \cdot \pscore({\scriptstyle T^n \to S^n}) + \\
    &  \hspace{20mm} \chart_{0, |\src|}[v, I^n]  \cdot\pscore({\scriptstyle T^n \to I^n}) \\
    & \chart[v, Root] = \sum_{n=1}^{|x|} \chart[v, T^n] \cdot \score({\scriptstyle Root \to T^n}). 
\vspace{-2mm}
\end{align*}
\paragraph{Cube pruning.}
Exact marginalization is still intractable  since there are theoretically infinitely many phrases that can be generated by a seq2seq model (e.g., there is  $\chart_{i:k}[v, S^1]$ for each possible phrase $v \in \Sigma^{+}$).
We follow the cube pruning strategy from \citet{huang-chiang-2005-better} and only store the top-$K$ target phrases 
within each chart item, and discard the rest.
During the bottom-up construction of the chart,  for  $A \in \{S, I\}$, $\chart_{i:k}[v, A^1]$ only stores the $K$ phrases 
generated by beam search with size $K$, and $\chart_{i:k}[v, A^n]$ constructs at most $2 \cdot (n - 1) \cdot K \cdot K$ target 
phrase translations and we only keep the top-$K$ phrases.
We call this decoding scheme \textbf{BTG-$N$ SeqSeq}. 
We generally found it to perform better than BTG-$1$ Seq2Seq but incur additional cost in decoding.\footnote{
Merging two sorted lists requires $O(K \log K)$ if implemented by priority queue~\citep{huang-chiang-2005-better}. 
Compared to the standard CKY algorithm, we also need to enumerate the number of segments $n$. Hence, 
the best time complexity of CKY decoding is $O\left((K \log K) |x|^4\right)$. In practice, we set a maximum number 
of segments $N$ and treat it as a hyperparameter to tune.} 
\input{tables/all_results.tex}

\vspace{-1mm}
\paragraph*{External Translation Rules.}
An important feature of CKY decoding is that we can constrain the prediction of BTG-Seq2Seq to incorporate new translation rules during inference. When we have access to new translation rules $\langle x_{i:k} \rightarrow v\rangle$ during inference (e.g., for idioms or proper nouns), we can enforce this by adding a rule $C[x_{i:k}] \to v$ and setting the score $\score(C[x_{i:k}] \to v) = 1$ (or some high value) during CKY decoding.

%% file: tables/all_results.tex
\begin{table*}
    \center
    \begin{adjustbox}{width=2.1\columnwidth,center}
    \begin{tabular}{ lcc c ccc ccccccc } 
        \toprule
        Task & \multicolumn{2}{c}{\textbf{SVO$\rightarrow$SOV}} & & \multicolumn{3}{c}{\textbf{EN$\rightarrow$ZH}} & & \multicolumn{3}{c}{\textbf{DE$\rightarrow$EN}} & & \multicolumn{2}{c}{\textbf{CHR$\leftrightarrow$EN}} \\
        
         Model & Novel-P & Few-Shot & & VP & NP & PP & & 500 & 1000 & 4978 & & CHR$\rightarrow$EN & EN$\rightarrow$CHR \\
        \midrule
        Seq2Seq & 1.2 & 22.1 & &  49.0 & 45.2 &	41.5 & &  22.9 &  25.7 &  29.7 & & 9.1 & 7.8 \\ 
        \hspace{2mm} + Trivial-Align & - & - & & 34.8 &	35.5 & 32.9 & & 17.8 &  19.1 &  23.0 & & 7.9 & 6.3 \\
        \hspace{2mm}  + Fast-Align  & - & - & & 42.0 & \bf{64.4} & 37.6 & & 21.1 &  24.0 &  29.1 & & 10.3 & 7.5 \\
        \midrule
        BTG-$1$ Seq2Seq & \bf{100} & 96.0 & & 55.9 &  60.8 & 48.5 & & 24.7  & 26.8 &  29.6 & & 11.2 & 9.3  \\
        BTG-$N$ Seq2Seq & \bf{100} & \bf{100} & & \bf{57.2} & 61.3 & \bf{49.8} & & \bf{25.1} &  \bf{29.0} &  \bf{31.0} & & \bf{13.6} & \bf{11.0} \\
        \bottomrule
       \end{tabular}
    \end{adjustbox}
    \vspace{-2mm}
    \caption{Accuracies on the toy SVO-SOV task, and BLEU scores on the few-shot English-Chinese (EN$\rightarrow$ZH) translation, low-data 
    German-English translation (DE$\rightarrow$EN), and the low-resource Cherokee-English (CHR$\leftrightarrow$EN) translation tasks. Trivial-Align and Fast-Align 
    refer to the baseline seq2seq models that are pre-trained with phrase pairs produced by the simplistic uniform-splitting heuristic and fast-align, respectively. For ZH$\rightarrow$EN, the columns (VP, NP, PP) denote different few-shot splits for pairing phrases with new contexts. For DE$\rightarrow$EN, the columns (500, 1000, 4978) refer to the number of training pairs. We use a pretrained mBART to initialize the seq2seq component for the DE$\rightarrow$EN experiments.}
    \label{tab:full_results}
    \vspace{-5mm}
\end{table*}

%% file: sections/experiment.tex
\vspace{-2mm}
\section{Experimental Setup}
\vspace{-2mm}

We apply BTG-Seq2Seq to a spectrum sequence-to-sequence learning tasks, starting from diagnostic toy datasets and then
gradually moving towards real machine translation tasks. Additional details (dataset statistics, model sizes, etc.) are given in appendix~\ref{app:statistics}.

\vspace{-2mm}
\paragraph*{Baselines.}
We use the standard Transformer-based seq2seq models~\citep{vaswani2017attention} as the backbone module for all experiments.
We consider three baselines: (1) a seq2seq model trained in a standard way, (2) a seq2seq model pretrained with phrase pairs extracted with an off-the-shelf tool, Fast-Align~\citep{dyer-etal-2013-simple},\footnote{
We extract phrase pairs that respect consistency constraints~\citep{koehn2009statistical} from Fast-Align word-level alignments.} and  (3)  a 
trivial baseline in which the model is pretrained on phrase pairs generated by randomly splitting a source-target sentence pair into  phrase pairs.  The latter two approaches study the importance of using learned alignments.

\vspace{-1mm}
\paragraph*{BTG-$1$ Seq2Seq and BTG-$N$ Seq2Seq.}
We focus on evaluating the two decoding modes of our model: CKY decoding where phrase-level translations are structurally combined 
to form a sentence-level prediction (BTG-$N$ Seq2Seq), and standard ``flat'' Seq2Seq decoding where sentence-level translations are generated directly by forcing $Root \rightarrow T^1$ (BTG-$1$ Seq2Seq).
In this way we test the effectiveness of  hierarchical alignments both as a structured translation model and as a regularizer on standard Seq2Seq learning.
BTG-$N$ Seq2Seq is sometimes referred simply as BTG-Seq2Seq in cases where CKY decoding is activated by default (e.g., constrained inference).

\paragraph{Backbone Transformers.} In each experiment, we use the same set of hyperparameters across models for fair comparison. One crucial hyperparameter for the backbone Transformers of BTG-Seq2Seq is to use relative positional encodings (as opposed to absolute ones) which are better at handling translation at phrase levels.Hence we  use relative positional encodings whenever possible except when utilizing pretrained Transformers that have been pretrained with absolute positional embeddings.

\vspace{-2mm}
\section{Main Results}
\vspace{-2mm}
\subsection{Toy SVO-SOV Translation}
\vspace{-1mm}
We start with a diagnostic translation task in which a model has to transform a synthetic sentence from SVO to SOV structure. 
The dataset is generated with the following synchronous grammar,
\begin{align*}
    & Root \to \langle \text{S} \ \text{V} \ \text{O} ,\, \ \text{S} \ \text{O} \ \text{V}\rangle, \\ & \hspace*{2mm} \text{S} \to  \text{NP}, \hspace*{2mm}
     \text{O} \to \text{NP}, \hspace*{2mm} \text{V} \to \text{VP}, \\
    & \text{NP} \to \langle np_s , \, \ np_t\rangle, \hspace*{2mm} \text{VP} \to \langle vp_s , \, \ vp_t\rangle,
\end{align*}
where the noun phrase pairs $\langle np_s, np_t \rangle$ and the verb phrase pairs $ \langle vp_s, \, vp_t \rangle$ are uniformly sampled from two nonoverlapping sets of synthetic phrase pairs with varying phrase lengths between 1 to 8. We create two generalization train-test splits.
\vspace{-1mm}
\paragraph*{Novel-Position.} For this split, a subset of $np_s, np_t$ phrase pairs are only observed in one syntactic position during training (i.e., only as a subject or only as an object). 
At test time, they appear in a new syntactic role. 
Each remaining $np_s, np_t$ pairs appears in both positions in training.
\vspace{-1mm}
\paragraph*{Few-Shot.}  In this setting, each subject (or object) phrase is observed in three examples in three different examples.  
During the evaluation, each subject phrase appears with context words  (i.e. a predicate and an object phrase) which are found in the training set but not with this subject phrase. 
\vspace{-1mm}
\paragraph*{Results.}
In Table~\ref{tab:full_results}, we observe that BTG-$N$ Seq2Seq  perfectly solves these two tasks.
We also observed that BTG-$N$ Seq2Seq correctly recovers 
the latent trees when $N=3$. Interestingly, even without structured CKY decoding, BTG-$1$ Seq2Seq
implicitly captures the underlying transduction rules, achieving near-perfect accuracy. In contrast,  standard seq2seq models fail in both settings.

\vspace{-1mm}
\subsection{English-Chinese: Phrasal Few-shot MT}
\vspace{-1mm}
Next, we test our models on the  English-Chinese translation dataset from \citet{li-etal-2021-compositional}, which was designed to test for compositional generalization in MT.
We use a subset of the original dataset and create 
three few-shot splits similar to the few-shot setting in the previous task. For example in the NP split noun phrases are paired with new contexts during evaluation.  Compared with the few-shot setting for the SVO-SOV task, this setup is more challenging due to a larger vocabulary and fewer training examples (approx. 800 data points).
\vspace{-1mm}
\paragraph*{Results.}
As shown in Table~\ref{tab:full_results}, both BTG-$1$ Seq2Seq and BTG-$N$ Seq2Seq improve on the standard Seq2Seq by a significant margin.  BTG-$N$ Seq2Seq also  outperforms the Fast-Align baseline\footnote{We pretrain on Fast-Align phrases and fine-tune on the original training set. This was found to perform better than jointly training on the combined dataset.} on VP and PP splits and slightly lags behind on NP, highlighting the benefits of jointly learning alignment along with the Seq2Seq model. 
\vspace{-1mm}
\paragraph{Analysis.}
In  Figure~\ref{fig:alignment} we visualize the  induced phrasal alignments for an example translation by
extracting the most probable derivation $\argmax_{d} p(d|x)$ for $n = 3$, which encodes the phrase-level correspondence
between $x$ and $y$.
In this example  ``at a restaurant'' is segmented and reordered before ``busy night'' according to the Chinese word order. 
We also give example translations in Table~\ref{tab:decode}, which shows  two failure modes of the standard seq2seq approach. The first failure mode is producing hallucinations. In the first example, the source phrase 
``hang out with the empty car he liked'' is translated to a Chinese phrase with a completely different meaning. This Chinese 
translation exactly matches a Chinese training example,  likely triggered by the shared word
\begin{CJK*}{UTF8}{gbsn}空车\end{CJK*} (``empty car''). The second failure mode is under-translation. In the second example,  standard seqsSeq completely
ignores the last phrase ``it was so dark'' in its Chinese translation, likely because  this phrase 
never co-occurs with the current context.
In contrast BTG-Seq2Seq is able to cover all source phrases, and in this case, predicts 
the corresponding Chinese phrase for ``it was so dark''.

\input{figures/alignment.tex}
\input{figures/decode.tex}
\input{figures/idiom.tex}
\vspace{-1mm}
\subsection{German-English: Low-Data MT}
\vspace{-1mm}
Next, we move onto more a realistic dataset and consider a low data German-English translation task , following~\citet{sennrich-zhang-2019-revisiting}.
The original dataset contains TED talks from the IWSLT 2014 DE-EN translation task~\citep{cettolo-etal-2014-report}.
Instead of training the seq2seq component from scratch, we use pre-trained mBART~\citep{liu-etal-2020-multilingual-denoising} to initialize the seq2seq component of our grammar (and the baseline seq2seq). This was found to perform much better than random initialization, and highlights an important feature of our approach which can work with arbitrary neural seq2seq models. (We explore this capability further in section~\ref{ssect:low}.)
The use of pre-trained mBART lets us consider extremely low-data settings, with 500 and 1000 sentence pairs, in addition to the 4978 setup from the original paper. We observe in Table~\ref{tab:full_results} that even the standard seq2seq achieves surprisingly reasonable BLEU scores in this low-data setting---much higher that scores of seq2seq models trained from scratch~\citep{sennrich-zhang-2019-revisiting}.\footnote{Their best seq2seq model with 4978 sentences obtains 16.57 BLEU, while phrase-based SMT obtains 15.87.}  BTG-$N$ Seq2Seq and BTG-$1$ Seq2Seq again outperform the baseline seq2seq model.  

Pretraining on Fast-Align harms performance, potentially due to it's being unable to induce meaningful phrase pairs in such low data regimes. In contrast, all our aligners (i.e., parsers) can also leverage pretrained features from mBART (we use the same mBART model to give span scores for all our parsers), which can potentially  improve structure induction and have a regularizing effect.
\input{tables/low-resource.tex}
\paragraph*{Injecting New Translation Rules.}
We investigate whether we can control the prediction of BTG-Seq2Seq by incorporating new translation rules during CKY inference.
In particular, we give new translation rules to mBART-based BTG-Seq2Seq for DE-EN translation. Since the model has  been trained only on 4978 pairs, the translations are far from perfect. But as shown in Table~\ref{tab:idiom} 
we observe that BTG-Seq2Seq can translate  proper nouns,  time expression, and even idioms to an extent. The idiom case is particularly interesting as pure seq2seq systems are known to be \emph{too} compositional when translating idioms  \cite{dankers2022can}. 

\vspace{-2mm}
\subsection{Cherokee-English: Low-Resource MT}
\vspace{-1mm}
\label{ssect:chr-en}
Our next experiments consider a true low-resource translation task with  Cherokee-English, using the dataset from \citet{zhang-etal-2020-chren}. This dataset contains around 12k/1k/1k examples for train/dev/test, respectively.
Unlike German-English, Cherokee was not part of the mBART pretraining set, and  thus we train all our models from scratch.  In Table~\ref{tab:full_results} we see that BTG-$N$ Seq2Seq and BTG-$1$ Seq2Seq again outperform the standard seq2seq.\footnote{However standard SMT is a strong baseline for Cherokee-English: \citet{zhang-etal-2020-chren} obtain 14.5 (Chr$\rightarrow$En) and 9.8 (En$\rightarrow$Chr) with Moses \citep{koehn-etal-2007-moses}.}

\subsection{Low Resource MT with Pretrained Models}
\label{ssect:low}
The current trend in low resource MT is to first pretrain a single multilingual model on a large number of language pairs \cite{tang2020multilingual,JMLR:v22:20-1307,costa2022no} and then optionally finetune it on specific language pairs. In our final experiment, we explore whether we can improve upon models that have already pretrained on bitext data. We focus mBART50-large which has been pretrained on massive monolingual \emph{and} bilingual corpora.  We randomly pick four low-resource languages from the ML50 benchmark \cite{tang2020multilingual}, each of which has fewer than 10$K$ bitext pairs. We also experiment with mT5-base \cite{xue-etal-2021-mt5} to see if we can plug-and-play other pretrained seq2seq models.

From Table~\ref{tab:low-mt}, we see that our BTG models outperform the standard finetuning baseline in four languages for mT5, 
and achieve slight improvement in two languages for mBART50.
We conjecture that this is potentially due to (1) mBART50's  being already trained on a large amount of bitext data and (2) mT5's use of relative positional encodings, which is more natural for translating at all phrase scales.  Finally, although we focus on pretrained seq2seq models in this work, an interesting extension would be to consider  \emph{prompt}-based hierarchical phrase-based MT where $p_{\text{seq2seq}} ( v | x_{i:k})$ is replaced with a large language model that has been appropriately prompted to translate phrases.\footnote{The source parser distribution could be set to uniform or learned on a small amount of labeled data.}

%% file: figures/alignment.tex
\begin{figure}[t]
    \begin{subfigure}[b]{0.495\textwidth}
        \centering
        \includegraphics[width=\linewidth]{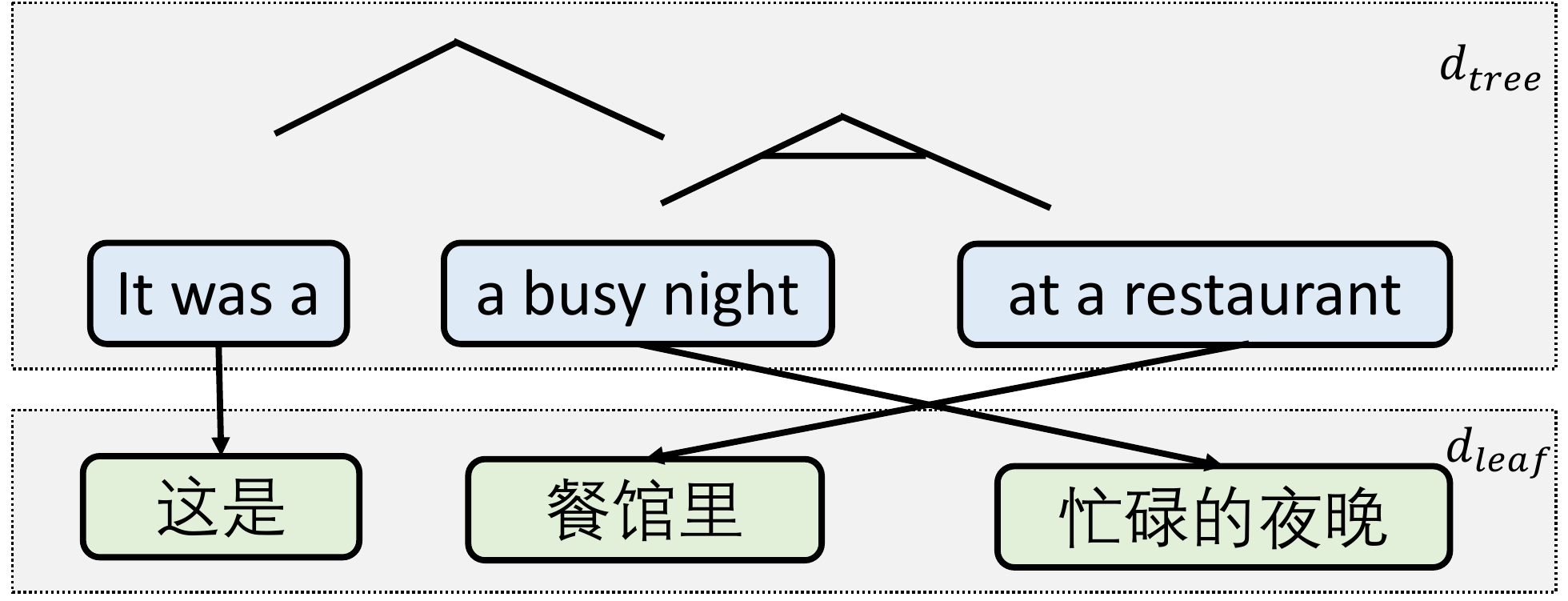}
    \end{subfigure}
        \vspace{-4mm}
    \caption{Induced hierarchical alignment by BTG-Seq2Seq on an English-Chinese example.}
    \vspace{-2mm}
    \label{fig:alignment}
\end{figure}

%% file: figures/decode.tex
\begin{CJK*}{UTF8}{gbsn}
\begin{table}[t]
    \center
    \resizebox{1.05 \columnwidth}{!}{%
    \vspace{-6mm}
   \begin{tabular}{rl}
    \toprule
    \textbf{Source}:  & she knew this but continued to \textit {hang out with the empty car he liked} . \\
     \textbf{Seq2Seq}: &  她 知道 这 一点 ， 但 还是 继续 \dashuline{他 喜欢 的 空 车 都 散发出 一股 难闻 的 气味 。} \\
    \textbf{BTG Seq2Seq}:  & 她 知道 这 一点 ， 但 继续 \dashuline{他 喜欢 的 空 车 在一起} 。 \\
      \textbf{Reference}: & 她 知道 这 一点 ， 但 继续 和 他 喜欢 的 空 车 在一起 。 \\
    \hline 
     \textbf{Source}: & he went inside the airplane on the floor \textit{and it was so dark} .  \\
    \textbf{Seq2Seq}:& 他进 了 地板 上 的 飞机 里 。  \\
     \textbf{BTG Seq2Seq}: & 他 走进 地板 上 的 飞机 \dashuline{里 面 一片 漆黑} 。 \\
    \textbf{Reference}: &  他 走进 地板 上 的 飞机 里 面 ， 里 面 一片 漆黑 。 \\
    \bottomrule
   \end{tabular} 
    }
         \vspace{-3mm}
   \caption{Chinese translations from a baseline seq2seq model and our BTG-Seq2Seq approach. Source phrases that are translated incorrectly by standard seq2seq are highlighted in italics, and target translations are delineated with dashed underlines.}
  \label{tab:decode}
  \vspace{-5mm}
\end{table}
\end{CJK*}

%% file: figures/idiom.tex
    \begin{table}[t]
        \center
        \resizebox{1.0 \columnwidth}{!}{%
       \begin{tabular}{rl}
        \toprule
        \textbf{Input}:  & Er hat das Unternehmen \textit{von Grund auf} aufgebaut  \\
        \textbf{\color{blue} New translation rule}: & {\color{blue} von Grund auf $\rightarrow$ from scratch} \\
        \textbf{Original Prediction}: & He built the company from the ground up  \\
        \textbf{New Prediction}: & He built the company \dashuline{from scratch} \\
        \textbf{Reference}: & He built the company from scratch  \\
        \textbf{Google Translate}: & He built the company from the ground up \\
        \hline
        \textbf{Input}: & Die europ\"aische Krise \textit{schließt den Kreis}   \\
         \textbf{\color{blue} New translation rule}: & {\color{blue} schließt den Kreis $\rightarrow$ is coming full circle}        \\
        \textbf{Original Prediction}: & The euro crisis closes the loop  \\
     \textbf{New Prediction}: & The European crisis \dashuline{is coming full circle} \\
     \textbf{Reference}: & The European crisis is coming full circle \\
        \textbf{Google Translate}: & The European crisis closes the circle \\
        \hline
         \textbf{Input}:& Die ARD strahlte gestern um 20:15 ``Aus dem Nichts" aus  \\
     \textbf{\color{blue} New translation rule}:  & {\color{blue}``Aus dem Nichts" $\rightarrow$  ``In the Fade", 20:15 $\rightarrow$ 8.15 pm } \\
        \textbf{Original Prediction}: & The ARD aired yesterday at 20:15 ``Out of Nowhere" \\
     \textbf{New Prediction}: & The ARD aired yesterday \dashuline{8.15 pm ``In the Fade"}  \\
        \textbf{Reference}: &ARD broadcast “In the Fade” yesterday at 8.15 pm \\
        \textbf{Google Translate}: & Yesterday at 8:15 p.m., ARD broadcast "Out of Nowhere". \\
        \bottomrule
       \end{tabular} 
       }
            \vspace{-3mm}
       \caption{Experiments on injecting new translation rules during inference. Examples are with the mBART model finetuned on the IWSLT 2014 German-to-English dataset. We manually provide new translation rules to BTG-SeqSeq, and decode with the constrained CKY algorithm. Injected phrases in the new predictions are highlighted with dashed underlines. We also provide results from Google Translate.}
      \label{tab:idiom}
      \vspace{-5mm}
    \end{table}

%% file: tables/low-resource.tex
\begin{table*}
    \center
    \begin{adjustbox}{width=2 \columnwidth,center}
    \begin{tabular}{ l c c c c } 
        \toprule
        \textbf{Models}  & Mongolian (MN$\rightarrow$EN) & Marathi (MR$\rightarrow$EN) & Azerbaijani (AZ$\rightarrow$EN) & Bengali (BN$\rightarrow$EN) \\
       \midrule 
       Finetune mT5 & 8.6 & 8.5 & 10.2 & 8.4   \\
       \hdashline 
       BTG-$1$ mT5 & 10.7  & 9.6 & 11.3 & 9.6 \\ 
       BTG-$N$ mT5 & 9.4 & 9.5 & 11.5 & 10.3 \\
        \midrule
       Finetune mBART50 & \cellcolor{lightgray}  11.2 & \cellcolor{lightgray}   \bf{11.6} & \cellcolor{lightgray}  15.5 & \bf{13.6} \\
       \hdashline 
       BTG-$1$ mBART50 & 11.5 & 11.2 & 15.8 & 12.8 \\
       BTG-$N$ mBART50 & \bf{12.1} & 10.9 & \bf{15.9} & 12.8 \\
        \bottomrule
       \end{tabular}
    \end{adjustbox}
        \vspace{-3mm}
    \caption{BLEU scores on four low-resource machine translation tasks. The backbone seq2seq models of our BTG models are 
    intialized with either mT5 \citep{xue-etal-2021-mt5} or mBART50 \citep{tang2020multilingual}.  Numbers in grey background are taken from \citet{tang2020multilingual}.}
    \label{tab:low-mt}
    \vspace{-5mm}
\end{table*}

%% file: sections/related_work.tex
\vspace{-2mm}
\section{Related Work}
\vspace{-2mm}
\paragraph*{Synchronous grammars}
Classic synchronous grammars and transducers have been widely explored in NLP for many applications \cite[\textit{inter alia}]{shieber-schabes-1990-synchronous,wu-1997-stochastic,eisner-2003-learning,ding-palmer-2005-machine,Nesson06inductionof,huang-etal-2006-syntax,wong-mooney-2007-learning,wang-etal-2007-jeopardy,graehl-etal-2008-training,blunsom-etal-2008-discriminative,blunsom2009scfg,cohn2009compression}.
In this work, we focus on the formalism of bracketing transduction grammars, which have been used for 
modeling reorderings in SMT systems \cite{nakagawa-2015-efficient,neubig-etal-2012-inducing}. 
Recent work has explored the coupling bracketing transduction grammars with
 with neural parameterization as a way to inject structural biases into neural sequence models \cite{wang2021structured}. 
Our work is closely related to the neural QCFG \cite{kim2021qcfg}, a neural parameterization of a quasi-synchronous 
grammars \cite{smith-eisner-2006-quasi}, which also defines a monolingual grammar conditioned on the source. However they do not experiment with embedding a neural seq2seq model within their grammar.
More generally our approach extends the recent line of work on neural parameterizations of classic grammars  \cite[\textit{inter alia}]{jiang-etal-2016-unsupervised,han2017dependency,han2019enhancing,kim2019pcfg,jin2019unsupervised,zhu2020lexpcfg,yang2021pcfg,yang2021lexical,zhao2020pcfg}, although unlike in these works we focus on the transduction setting.

\vspace{-2mm}
\paragraph*{Data Augmentation}
Our work is also related to the line of work on utilizing grammatical or alignment structures to guide flexible neural seq2seq models via data augmentation \cite{jia-liang-2016-data,fadaee2017data,andreas-2020-good,akyurek2021learning,shi-etal-2021-substructure,yang-etal-2022-subs,qiu-etal-2022-improving} 
or auxiliary supervision  \cite{cohn-etal-2016-incorporating,mi-etal-2016-supervised,liu-etal-2016-neural,yin-etal-2021-compositional}. 
In contrast to these works our data augmentation module has stronger inductive biases for hierarchical structure due to explicit use of latent tree-based alignments.

%% file: sections/conclusion.tex
\vspace{-1mm}
\section{Conclusion}
\vspace{-1mm}
We presented a neural transducer that maintains the flexibility and seq2seq models but  still incorporates hierarchical phrases both as a source of inductive bias during training and as explicit constraints during inference. We formalized our model as a synchronous grammar and developed an efficient variational inference algorithm for training.
Our model performs well compared to baselines on small MT benchmarks both as a regularized seq2seq model and as a synchronous grammar.

%% file: sections/limitations.tex
\section{Limitations}

Our work has several  limitations. While variational inference enables more efficient training than full marginalization ($O(L^3)$ vs. $O(L^7)$), it is still much more expensive compute- and memory-wise than regular seq2seq training. This currently precludes our approach as a viable alternative on large-scale machine translation benchmarks, though advances in GPU-optimized dynamic programming algorithms could improve  efficiency  \cite{alex2020torchstruct}. (However, as we show in our experiments it is possible to finetune a pretrained seq2seq MT system using our approach.) Cube-pruned CKY decoding is also much more expensive than regular seq2seq decoding. In general, grammar- and automaton-based models enable a greater degree of model introspection and interpretability. However our use of black-box seq2seq systems within the grammar still makes our approach not as interpretable as classic transducers.

\section*{Acknowledgements}

This study was supported by MIT-IBM Watson AI Lab.

%% file: sections/appendix.tex
\section{Parameterization of BTG-Seq2Seq }
\label{app:parameterization}
The basic strategy for our parameterization  is to take 
as much advantage of the backbone Transformer seq2seq models as possible, 
which is the main source of rich neural representations (especially when working with pretrained models).

\paragraph*{Seq2Seq $\boldsymbol{p(\dleaf| \src, \dtree)}$.}

The leaf nodes of $\dtree$ are the production rules of $C[x_{i:k}] \rightarrow v$,
which encode the correspondence between source segment $x_{i:k}$ and target segment $v$. 
The probability of such a rule is modeled by a conventional Transformer-based seq2seq model~\citep{vaswani2017attention}.
In BTG-Seq2Seq, the underlying seq2seq is expected to handle segments at all levels, including sentence- and phrase-level translations.
We found it beneficial to use two sets of special tokens to indicate the beginning/end of sentences, one for sentence-level translations and the other for phrase-level translations. Concretely, in addition to the typical `BOS' and `EOS' which are used to mark the beginning and the end 
of a sentence, we additionally use `Seg-BOS' and `Seg-EOS' to mark the beginning and the end of a segment. 

\paragraph*{Source parser $\boldsymbol{p(\dtree|\src,n)}$.} To obtain a probabilistic parser, we need to assign a score to each rule involved in $\dtree$. For the following rules,
\begin{align*}
&  \forall n \in \{1, 2 \dots |x|\}, \\
&  \hspace{5mm}  T^n \to S^n[x], \hspace{4mm} T^n \to I^n[x],
\end{align*}
we assign them a  weight of 1 for simplicity, as we find that using trainable weights does not help during our initial experiments. 

We use neural features to parameterize the score of the following rules:
\begin{align*}
& \forall i,j,k \in \{0, 1 \dots |x|\} && \textrm{s.t.} \ \  i < j < k \\
    & \hspace{5mm} S^n[x_{i: k}] \to I^l[x_{i: j}] S^r[x_{j: k}] \\
    & \hspace{5mm} S^n[x_{i: k}] \to I^l[x_{i: j}] I^r[x_{j: k}] \\
    & \hspace{5mm} I^n[x_{i: k}] \to S^l[x_{j: k}] I^r[x_{i: j}] \\ 
    & \hspace{5mm} I^n[x_{i: k}] \to S^l[x_{j: k}] S^r[x_{i: j}].
\end{align*}
For example the score
\begin{align*}
\score({\scriptstyle S^n[x_{i: k}] \to I^l[x_{i: j}] S^r[x_{j: k}]}) = \exp \left( e_{S}^\top f(\mathbf{\tilde h}_{i:j}, \mathbf{\tilde h}_{j:k}) \right) 
\end{align*}
relies on the difference feature $\mathbf{\tilde h}_{i:j}$ and $\mathbf{\tilde h}_{j:k}$, following \citet{kitaev2018parsing}. 
In our parameterization, rules that have the same right-hand spans will have the same score regardless 
of their nonterminals. That is, the score of $I^n[x_{i: k}] \to S^l[x_{j: k}] I^r[x_{i: j}]$
is identical to $I^n[x_{i: k}] \to S^l[x_{j: k}] S^r[x_{i: j}]$.
These difference features are calculated using 
\begin{align*}
\mathbf{\tilde h}_{i:j} = \text{Concat}[\overrightarrow{ \mathbf{h}}_j - \overrightarrow{\mathbf{h}}_i; \overleftarrow{ \mathbf{h}}_{j+1} - \overleftarrow{\mathbf{h}}_{i+1}]
\end{align*}
where $\overrightarrow{\mathbf{h}}_i$ and $\overleftarrow{\mathbf{h}}_i$ are the ``forward'' and ``backward'' representations from \citet{kitaev2018parsing}.\footnote{Originally, the forward and backward representations were defined based on the forward/backward hidden states from a bidirectional LSTM~\citep{stern-etal-2017-minimal}. \citet{kitaev2018parsing} extend it to the Transformer architecture by  splitting the hidden representation $\mathbf{h}_i$ from the Transformer encoder into halves and treating them as the forward and backward representations, i.e., $\mathbf{h}_i = \text{Concat}[\overrightarrow{\mathbf{h}}_i ; \overleftarrow{\mathbf{h}}_i]$.}

The score of a derivation tree is the sum of the scores of the production rules that form the tree, and we normalize 
the score to obtain the distribution over all possible derivations,
\begin{equation*}
    p(\dtree|\src, n)  =  \left( \prod_{R \in \dtree} \score(R) \right)  / Z_{\text{tree}}(\src, n),
\end{equation*}
where $Z_{\dtree}(\src, n) = \sum_{\dtree} \left( \prod_{R \in \dtree} \score(R)\right)  $ is the partition function. The dynamic program for computing the partition function is given by Algorithm~\ref{algo:inside-tree}.

\vspace{-1mm}
\paragraph*{Variational parser $\boldsymbol{q(\dtree, \dleaf|\src,\tgt,n)}$.}
We factorize the variational synchronous parser as
\begin{align*}
    q(\dtree, \dleaf|\src,\tgt,n) &= q(\dtree|\src,\tgt,n) \times \\ &\,\,\,\, \,\, \, q(\dleaf|\src,\tgt, \dtree),
\end{align*}
which makes it possible to sample hierarchical alignments in $O(L^3)$ where $L=\min(|x|, |y|)$.
The parameterization of $q(\dtree|\src,\tgt,n)$ is very similar to the parameterization of $p(\dtree|\src,n)$, except that $\mathbf{h}_i$ (which are used to compute the difference features) are the hidden states of a Transformer \emph{decoder}, with the backbone Transformer seq2seq running from $\tgt$ to $\src$. This choice makes it possible to condition the contextualized representations of $\src$ on the target $\tgt$. Calculating the partition function uses the same dynamic program as in the prior parser (i.e., Algorithms~\ref{algo:inside-tree}). Algorithm~\ref{algo:sample-tree} shows the dynamic program for sampling $\dtree$.

Given $\src, \tgt$ and $\dtree$, the only rules that remain unknown for the variational parser are the target 
segments $y_{a:c}$ in $C[x_{i:k}] \rightarrow y_{a:c}$. Thus $q(\dleaf|\src,\tgt, \dtree)$ is effectively a hierarchical segmentation model. We use a grammar with the following rules,
\begin{align*}
& Root \to T^n[y], \ 
T^m[y_{a:c}] \to T^l[y_{a:b}] T^r[y_{b:c}],
\end{align*}
to model the segmentations. Since the segmentation is conditioned on $\dtree$, we need to ensure that the tree structures of $\dleaf$ and $\dtree$ is identical. (Note that by conditioning on $\dtree$,  we only utilize the tree topology of $\dtree$ for efficiency.)

We use the following function to score the rules:
\begin{equation*}
    \score({\scriptstyle [T^m[y_{a:c}] \to T^l[y_{a:b}] T^r[y_{b:c}] }) = \exp \left( f(\tilde{ \mathbf{h}}_{a:b}, \tilde{ \mathbf{h}}_{b:c}) \right),
\end{equation*}
where $\tilde{ \mathbf{h}}_{a:b}$, $\tilde{ \mathbf{h}}_{b:c}$ are the difference features based on the contextualized representations of $\tgt$, and 
$f$ is an MLP that emits a scalar score.
We can then obtain a  distribution over $\dleaf$ via,
\begin{align*}
  &  q(\dleaf|\src, n, \dtree) \hspace{-1mm} = \hspace{-1mm}  \left( \prod_{R \in \dleaf} \score(R)  \right) / Z_{\text{leaf}}(\dtree, \tgt)
\end{align*}
where $Z_{\dleaf}(\dtree, \tgt) = \sum_{\dleaf} \left( \prod_{R \in \dleaf} \score(R)  \right)$ is the partition function. Algorithm~\ref{algo:inside-leaf} shows the dynamic program for calculating the partition function above, and Algorithm~\ref{algo:sample-leaf} gives the dynamic program for sampling from $q(\dleaf|\src, n, \dtree)$. These dynamic programs modify the standard inside algorithm to only sum over chart entries that are valid under $\dtree$.

Since we share the encoder and decoder across all components (i.e., prior/variational parsers and the seq2seq model), the only additional parameters for BTG-Seq2Seq are the parameters of the MLP layers for the three CRF parsers.

\input{tables/hyper-1.tex}
\input{tables/hyper-2.tex}
\vspace{-1mm}
\paragraph*{Number of segments  $\boldsymbol {p(n|x)$, $q(n|x, y)}$.}

We use a truncated geometric distribution of the following form,
\begin{equation*} 
    p(n) =
    \begin{cases}
     & \lambda (1 - \lambda)^{n - 1}, \  n \in \{1, 2, \dots N-1\} \\
     & (1 - \lambda)^{n - 1}, \ n = N
    \end{cases}
\end{equation*}
where $N$ is the maximum number of segments. For the prior  $p(n|x)$, we have $N = |x|$. For the variational posterior $q(n|x, y)$, we have
$N = \min(|x|, |y|)$. In practice, we set an additional upper bound (chosen from $[3, 4 \dots 8]$) for efficient training. $\lambda$ is chosen from $\{0.5, 0.6, 0.7, 0.8, 0.9\}$. %

\vspace{-2mm}
\section{Evidence Lower Bound}
\vspace{-2mm}
\label{app:elbo}
Figure~\ref{fig:elbo} shows the lower bound derivation.
\begin{figure*}
\begin{align*}
 \log p(\tgt|\src)  &  =  \log \sum_{d \in D(\src, \tgt)} p(d|\src) =  \log \sum_{d \in D(\src, \tgt)} p(n|\src) p(\dtree | \src, n) p(\dleaf | \src, \dtree) \\ 
  & \geq \E_{q(n | \src, \tgt)}\Big\{ \boxed{ \log \sum_{\dtree, \dleaf} p(\dtree | \src, n) p(\dleaf | \src, \dtree) } \Big\} - \KL[q(n|\src, \tgt) \Vert p(n|\src)]   \\
  & { \quad \quad \quad \geq \E_{q(\dtree | \src, \tgt, n)}\Big\{ \boxed{ \log  \sum_{\dleaf} p(\dleaf | \src, \dtree) }  - \KL[q(\dtree | \src, \tgt, n) \Vert p(\dtree|\src, n)] \Big  \}  } \\
  & {\quad \quad  \quad \quad \quad \quad   \geq \E_{q(\dleaf | x, y, \dtree)}[\log  p(\dleaf | \src, \dtree)] + \ENT[q(\dleaf | x, y, \dtree)]  }  \\
  \\
  \log p(\tgt|\src) & 
        \geq \E_{ q(n | \src, \tgt) } \Big \{ \E_{q(\dtree | \src, \tgt, n)}   \Big[  
          \E_{q(\dleaf | x, y, \dtree)}[\log  p(\dleaf | \src, \dtree)] + \ENT[q(\dleaf | x, y, \dtree)]\Big]  \\ &
      - \KL [q(\dtree | \src, \tgt, n) \Vert p(\dtree | \src, n) ] \Big \} - \KL[q(n|\src, \tgt) \Vert p(n|\src)]
\end{align*}
\vspace{-5mm}
\caption{Derivation of the evidence lower bound used for training. (Top) Here, line 3 lower bounds the boxed term in line 2, while line 4 lower bounds the boxed term in line 3. (Bottom) Taking these steps together, we can obtain the final lower bound.}
\label{fig:elbo}
\vspace{-2mm}
\end{figure*}

\vspace{-2mm}
\section{Dynamic Programs for the Parsers}
\vspace{-2mm}
\label{app:algo}

\subsection*{Sampling Algorithms}
\vspace{-1mm}
The algorithms for sampling from $p(\dtree| \src, n)$ and $q(\dleaf | \src, \dtree)$ are provided in Algorithm~\ref{algo:sample-tree} and \ref{algo:sample-leaf}. Sampling from $q(\dtree| \src, \tgt, n)$ is omitted, as it is the same algorithm as Algorithm~\ref{algo:sample-tree}. 

\vspace{-1mm}
\subsection*{Computing the Entropy and KL}
\vspace{-1mm}
We provide the dynamic program to compute $\ENT[q(\dleaf | x, y, \dtree)] $ in Algorithm~\ref{algo:leaf-entropy}, and $\KL [q(\dtree | \src, \tgt, n) \, || \, p(\dtree | \src, n) ]$ in Algorithm~\ref{algo:tree-kl}. Note that $\KL[q(n|\src, \tgt) \Vert \, p(n | \src)]$ is a constant, and we can ignore it during optimization.
In practice the algorithms are implemented in log space for numerical stability.
\vspace{-1mm}
\subsection*{Gradient Estimation}
\vspace{-1mm}
We use policy gradients~\citep{williams1992simple} to optimize the lower bound. Two techniques are additionally 
employed to reduce variance. First, we use a self-critical control variable~\citep{rennie2017self} where the scores from argmax trees 
are used as a control variates. Second, we utilize a sum-and-sample strategy~\citep{liu2019rao} where we always train on $n=1$, i.e., sentence-level translations, and sample another $n > 1$ to explore segment-level translations.
\vspace{-1mm}
\section{Recipe for Using BTG-Seq2Seq}
\vspace{-2mm}

Although it is expensive to train our BTG-Seq2Seq from scratch due to $O(L^3)$ time complexity, we find that a pretrain-then-finetune strategy is a  practical way 
to use BTG-Seq2Seq. During the pretraining stage, we train a Transformer seq2seq in the standard way until convergence, or simply use public model checkpoints. Then during the finetuning stage, we use the pretrained model as the backbone seq2seq for our BTG-SeqSeq. Usually we use a relatively smaller learning rate to finetune the backbone seq2seq during finetuning.

\section{Dataset Statistics and Hyperparameters}

\label{app:statistics}
We show the statistics of the datasets and the hyperparameters of the model in Tables~\ref{tab:hyper-1} and \ref{tab:hyper-2}.

\input{algo/inside_tree.tex}
\input{algo/sample_tree.tex}
\input{algo/inside_leaf.tex}
\input{algo/sample_leaf.tex}
\input{algo/entropy.tex}
\input{algo/kl.tex}

%% file: tables/hyper-1.tex
\begin{table*}[t]
    \center
    \begin{adjustbox}{width=2.1\columnwidth,center}
    \begin{tabular}{ l cc cccccccc } 
        \toprule
                    & \multicolumn{2}{c}{\textbf{SVO$\rightarrow$SOV}} & \multicolumn{3}{c}{\textbf{EN$\rightarrow$ZH}} & \multicolumn{3}{c}{\textbf{DE$\rightarrow$EN}} & \multicolumn{2}{c}{\textbf{CHR$\leftrightarrow$EN}} \\
         & Novel-P & Few-Shot & VP & NP&  PP & 500 & 1000 & 4978 & chr-en & en-chr \\
         \hline
        Train & 20000 & 2000 &  914 & 812 & 906 & 500 & 1000 & 4978 & 11638 & 11638 \\
        Dev& 5000 & 500 & 200 & 200 & 200 & 7584 & 7584 & 7584 & 1000 & 1000\\
        Test & 5000 & 500 & 200 & 200 & 200 & 6759 & 6759 & 6759 & 1000 & 1000\\
        \midrule
        Transformer Encoder Layer  & 2 & 2 &  6 & 6 & 6	& 12 &  12 &  12 & 6  & 6\\ 
        Transformer Encoder Size  & 1024 & 1024 &  512 & 512 & 512 &  1024 &  1024 & 1024 & 512 & 512 \\ 
        Transformer Decoder Layer  & 2 & 2 &  6 & 6 & 6 &  12 &  12 & 12 & 6 & 6 \\ 
        Transformer attention head  & 8 & 8 &  8 & 8 &	8 &  16 &  16 & 16 & 8 & 8 \\ 
        Positional embedding type   & relative & relative &  relative & relative &	relative &  absolute &  absolute &  absolute & relative & relative \\ 
        \bottomrule
       \end{tabular}
    \end{adjustbox}
    \vspace{-2mm}
    \caption{Dataset statistics and model hyperparameters used for experiments. Apart from the standard absolute positional encodings used by mBART, we use relative positional encodings whenever possible.}
    \label{tab:hyper-1}
    \vspace{-2mm}
\end{table*}

%% file: tables/hyper-2.tex
\begin{table}[t]
    \center
    \begin{adjustbox}{width=\columnwidth,center}
    \begin{tabular}{ l cc ccc ccc cc } 
        \toprule
         & \multicolumn{1}{c}{\textbf{Mongolian (MN)}} & \multicolumn{1}{c}{\textbf{Marathi (MR)}} & \multicolumn{1}{c}{\textbf{Azerbaijani (AZ)}} & \multicolumn{1}{c}{\textbf{Bengali (BN)}} \\
         \hline
        Train & 7607 & 9840 &  5946 & 4649 \\
        Dev & 372 & 767 & 671 & 896 \\
        Test & 414 & 1090 & 903 & 216 \\
        \bottomrule
       \end{tabular}
    \end{adjustbox}
    \vspace{-3mm}
    \caption{Dataset sizes for four low-resource languages we choose from ML50 benchmark~\citep{tang2020multilingual}.}
    \label{tab:hyper-2}
\end{table}

%% file: algo/inside_tree.tex
\begin{algorithm*}[ht]
    \caption{Inside algorithm for $p(\dtree|\src, n)$. Rules are highlighted in {\color{blue} blue} }
     \hspace*{3mm} \textbf{Input:} $x$: source sentence, $n$: number of segments 
    \begin{algorithmic}
    \Function{Inside-Tree}{$x, n$}
        \State Initialize $\beta_{\dots}[S^1] = 0$, $\beta_{\dots}[I^1] = 0$
        \For{$m = 2$ \textbf{to} $n$}  \Comment number of segments
            \For{$w = 1$ \textbf{to} $|x|$} \Comment width of spans
                \For{$i = 0$ \textbf{to} $|x| - w$}  \Comment start point
                    \State $k = i + w $ \Comment end point
                    \For{$j = i + 1$ \textbf{to} $k-1$}
                        \For{$l = 1$ \textbf{to} $m-1$} \Comment  number of segments in the left part
                    \State $r = m - l$ \Comment  number of segments in the right part
                            \State $\beta_{i:k}[S^m] \mathrel{+}= \score({\scriptstyle \color{blue} S^m[x_{i:k}] \to I^l[x_{i:j}] S^r[x_{j:k}]}) \cdot \beta_{i:j}[I^l] \cdot \beta_{j:k}[S^r]  $
                            \State $\beta_{i:k}[S^m] \mathrel{+}=\score({\scriptstyle \color{blue} S^m[x_{i:k}] \to I^l[x_{i:j}] I^r[x_{j:k}]}) \cdot \beta_{i:j}[I^l] \cdot \beta_{j:k}[I^r]  $
                            \State $\beta_{i:k}[I^m] \mathrel{+}= \score({\scriptstyle \color{blue} I^m[x_{i:k}] \to S^l[x_{j:k}] I^r[x_{i:j}]}) \cdot \beta_{j:k}[S^l] \cdot \beta_{i:j}[I^r]  $
                            \State $\beta_{i:k}[I^m] \mathrel{+}= \score({\scriptstyle \color{blue} I^m[x_{i:k}] \to S^l[x_{j:k}] S^r[x_{i:j}]}) \cdot \beta_{j:k}[S^l] \cdot \beta_{i:j}[S^r]  $
                        \EndFor
                    \EndFor
                \EndFor
            \EndFor
        \EndFor
        \LineComment{Compute locally-normalized scores $\pscore$}
        \For{$m = 2$ \textbf{to} $n$}  \Comment number of segments
            \For{$w = 1$ \textbf{to} $|x|$} \Comment width of spans
                \For{$i = 0$ \textbf{to} $|x| - w$}  \Comment start point
                    \State $k = i + w $ \Comment end point
                    \For{$j = i + 1$ \textbf{to} $k-1$}
                        \For{$l = 1$ \textbf{to} $m-1$} \Comment  number of segments in the left part
                    \State $r = m - l$ \Comment  number of segments in the right part
                            \LineComment{Denote $\scriptstyle \color{blue} S^m[x_{i:k}] \to I^l[x_{i:j}] S^r[x_{j:k}]$ as ${\color{blue} R_1}$}
                            \State $ \pscore({\color{blue} R_1}) = \left( \score({\color{blue} R_1}) \cdot \beta_{i:j}[I^l] \cdot \beta_{j:k}[S^r] \right) / \beta_{i:k}[S^m] $
                             \LineComment{Denote $\scriptstyle \color{blue} S^m[x_{i:k}] \to I^l[x_{i:j}] I^r[x_{j:k}]$ as ${\color{blue} R_2}$}
                            \State $ \pscore({\color{blue} R_2}) = \left( \score({\color{blue} R_2}) \cdot \beta_{i:j}[I^l] \cdot \beta_{j:k}[I^r] \right) / \beta_{i:k}[S^m]  $
                             \LineComment{Denote $\scriptstyle \color{blue} I^m[x_{i:k}] \to S^l[x_{j:k}] I^r[x_{i:j}]$ as ${\color{blue} R_3}$}
                            \State $\pscore({\color{blue} R_3})= \left( \score({\color{blue} R_3}) \cdot \beta_{j:k}[S^l] \cdot \beta_{i:j}[I^r] \right) / \beta_{i:k}[I^m]  $
                             \LineComment{Denote $\scriptstyle \color{blue} I^m[x_{i:k}] \to S^l[x_{j:k}] S^r[x_{i:j}]$ as ${\color{blue} R_4}$}
                            \State $\pscore({\color{blue} R_4})= \left( \score({\color{blue} R_4}) \cdot \beta_{j:k}[S^l] \cdot \beta_{i:j}[S^r] \right) / \beta_{i:k}[I^m]  $
                        \EndFor
                    \EndFor
                \EndFor
            \EndFor
        \EndFor
        \LineComment{$T$ rules are assigned a trivial score: $\score({\scriptstyle \color{blue} T^n \to S^n[\src]})= 1, \quad \score({\scriptstyle \color{blue} T^n \to I^n[\src]})= 1$}
        \State $\beta[T^n] = \beta_{0: |x|}[S^n] + \beta_{0: |x|}[I^n] $, $\quad$ $Z_{\text{tree}}(\src, n) = \beta[T^n]$ \Comment{partition function}
        \State $\pscore({\scriptstyle \color{blue} T^n \to S^n[\src]}) = \beta_{0: |x|}[S^n] / \beta[T^n] $, $\quad$ $\pscore({\scriptstyle \color{blue} T^n \to I^n[\src]}) = \beta_{0: |x|}[I^n] / \beta[T^n] $
        \State \Return $Z_{\text{tree}}(\src, n), \beta, \pscore $
    \EndFunction
    \end{algorithmic} 
    \label{algo:inside-tree}
\end{algorithm*}

%% file: algo/sample_tree.tex
\begin{algorithm*}[ht]
    \caption{Top-down sampling $\dtree$ from $p(\dtree|x, n)$. Rules are highlighted in {\color{blue} blue}}
     \hspace*{3mm} \textbf{Input:} $\pscore$: normalized scores obtained from \Call{Inside-Tree}{$\cdot$}, $n$: number of segments
    \begin{algorithmic}
    \Function{Sample-Tree}{$\pscore, n$}
        \State Initialize an empty tree $\dtree$
        \State Sample $A \in \{S, I\}$ wrt. $\pscore({\scriptstyle \color{blue} T^n \to A^n[\src]})$
        \State \Call{Recur-Sample-Tree}{$0$, $|x|$, $n$, $A$, $\pscore$, $\dtree$}
        \State \Return $\dtree$
    \EndFunction
    \texttt{\\}
    \LineComment{Arguments: $i$: start point, $k$: end point, $n$: number of segments, $A$: nonterminal)}
    \Function{Recur-Sample-Tree}{$i$, $k$, $n$, $A$,  $\pscore$, $\dtree$}
        \If{$n=1$}
        \State Add rule ${\scriptstyle \color{blue} A^1[x_{i:k}] \to C[x_{i:k}]}$ to $\dtree$
        \State \Return
        \EndIf
        \If{$A = S$}
            \LineComment{$1 \leq l < n,  A,B \in \{S, I\}, i < j < k$ are the random variables, $r = n - l$}
            \State Sample rule ${\scriptstyle \color{blue} S^n[x_{i:k}] \to A^l[x_{i:j}] B^r[x_{j:k}]}$ according to $\pscore(\cdot)$
            \State Add ${\scriptstyle \color{blue} S^n[x_{i:k}] \to A^l[x_{i:j}] B^r[x_{j:k}]}$ to $\dtree$
            \State \Call{Recur-Sample-Tree}{$i$, $j$, $l$, $A$,  $\pscore$, $\dtree$}
            \State \Call{Recur-Sample-Tree}{$j$, $k$, $r$, $B$, $\pscore$, $\dtree$}
        \Else \Comment{$A = I$}
            \LineComment{$1 \leq l < n,  A, B \in \{S, I\}, i < j < k$ are the random variables, $r = n - l$}
            \State Sample rule ${\scriptstyle \color{blue} I^n[x_{i:k}] \to A^l[x_{j:k}] B^r[x_{i:j}]}$ according to $\pscore(\cdot)$
            \State Add ${\scriptstyle \color{blue} I^n[x_{i:k}] \to A^l[x_{j:k}] B^r[x_{i:j}]}$ to $\dtree$
            \State \Call{Recur-Sample-Tree}{$j$, $k$, $l$, $A$, $\pscore$, $\dtree$}
            \State \Call{Recur-Sample-Tree}{$i$, $j$, $r$, $B$, $\pscore$, $\dtree$}
        \EndIf
    \EndFunction
    \end{algorithmic} 
    \label{algo:sample-tree}
\end{algorithm*}

%% file: algo/inside_leaf.tex
\begin{algorithm*}[ht]
    \caption{Inside algorithm for $q(\dleaf| \src, \tgt, \dtree)$.}
     \hspace*{3mm} \textbf{Input:} $\tgt$: target sentence, $\dtree$: derivation tree
    \begin{algorithmic}
    \Function{Inside-Leaf}{$\tgt, \dtree$}
        \State Infer number of segments $n$ from $\dtree$
        \State Initialize $\beta_{\dots}[\dots] = 0$
        \For{$m = 2$ \textbf{to} $N$}  \Comment number of segments
            \For{$w = 1$ \textbf{to} $|x|$} \Comment width of spans
                \For{$i = 0$ \textbf{to} $|x| - w$}  \Comment start point
                    \State $k = i + w $ \Comment end point
                    \For{$j = i + 1$ \textbf{to} $k-1$}
                        \For{$l = 1$ \textbf{to} $m-1$} \Comment  number of segments in the left part
                        \State $r = m - l$ \Comment  number of segments in the right part
                            \If{there exist a rule like $A^m \to B^l C^r$ in $\dtree$}
                                \State $\beta_{i:k}[T^m] \mathrel{+}= \score({\scriptstyle \color{blue} T^m[y_{i:k}]  \to T^l[y_{i:j}] T^r[y_{j:k}]}) \cdot \beta_{i:j}[T^l] \cdot \beta_{j:k}[T^r]  $
                            \EndIf
                        \EndFor
                    \EndFor
                \EndFor
            \EndFor
        \EndFor
        \State \Return $\beta$
    \EndFunction
    \end{algorithmic} 
    \label{algo:inside-leaf}
\end{algorithm*}

%% file: algo/sample_leaf.tex
\begin{algorithm*}[ht]
    \caption{Top-down sampling $d_{\text{leaf}}$ from $q(d_{\text{leaf}}|\src, \tgt, d_{\text{tree}})$.}
     \hspace*{3mm} \textbf{Input:} $\beta$: inside scores returned from \Call{Inside-Leaf}{$\cdot$},  $\dtree$: derivation tree, $y$: target sentence
    \begin{algorithmic}
    \Function{Sample-Leaf}{$\beta$, $\dtree, y$}
        \State Infer the number of segments $n$ from $\dtree$
        \State Initialize an empty tree $\dleaf$
        \State \Call{Recur-Sample-Leaf}{$0$, $|y|$, $n$, $\beta$, $\dtree$, $\dleaf$}
        \State \Return $\dleaf$
    \EndFunction
    \texttt{\\}
    \LineComment{Arguments: $i$: start point, $k$: end point, $n$: number of segments}
    \Function{Recur-Sample-Leaf}{$i$, $k$, $n$, $\beta$, $\dleaf$, $\dtree$}
        \If{$n=1$}
            \State Add rule ${\scriptstyle \color{blue} T^1[y_{i: k}] \to y_{i:k}}$ to $d_{\text{leaf}}$ 
            \State \Return 
        \EndIf
        \texttt{\\}
        \State Infer $l, r$ based on $\dtree$ and $n$
        \State Compute $Z = \sum_{j} (\beta_{i:j}[T^l] \cdot \beta_{j:k}[T^r]) $  \Comment{normalization term}
        \For{$j = i + 1$ \textbf{to} $k-1$}
            \State $\pscore({\scriptstyle \color{blue} T^n[y_{i:k}] \to T^l[y_{i:j}] T^r[y_{j:k}]}) = (\beta_{i:j}[T^l] \cdot \beta_{j:k}[T^r]) / Z  $
        \EndFor
        \texttt{\\}
        \State Sample rule ${\scriptstyle \color{blue} T^n[y_{i:k}] \to T^l[y_{i:j}] T^r[y_{j:k}]}$, and add it to $\dleaf$  \Comment{the rule only encodes split point $j$}
        \State \Call{Recur-Sample-Leaf}{$i$, $j$, $l$, $\beta$, $\dtree$, $\dleaf$}
        \State \Call{Recur-Sample-Leaf}{$j$, $k$, $r$, $\beta$, $\dtree$, $\dleaf$}
    \EndFunction
    \end{algorithmic} 
    \label{algo:sample-leaf}
\end{algorithm*}

%% file: algo/entropy.tex
\begin{algorithm*}[ht]
    \caption{Calculating the entropy of $\ENT[ q(\dleaf| \src, \tgt, \dtree)]$}
     \hspace*{3mm} \textbf{Input:} $\beta$: inside scores returned by \Call{Inside-Leaf}{$\cdot$}, $\dtree$: derivation tree, $y$: target sentence
    \begin{algorithmic}
    \Function{Compute-Leaf-Entropy}{$\beta$, $\dtree, y$}
        \State Find the number of segments $n$ from $\dtree$
        \State \Return \Call{Recur-Compute-Leaf-Entropy}{$0$, $|y|$, $n$, $\beta$, $\dtree$}
    \EndFunction
    \texttt{\\}
    \LineComment{Arguments: $i$: start point, $k$: end point, $n$: number of segments}
    \Function{Recur-Compute-Leaf-Entropy}{$i$, $k$, $n$, $\beta$, $\dtree$}
        \If{$n=1$}
            \State \Return 0 
        \EndIf
        \texttt{\\}
        \State Infer $l, r$ based on $\dtree$ and $n$
        \State Compute $Z = \sum_{j} (\beta_{i:j}[T^l] \cdot \beta_{j:k}[T^r]) $  \Comment{normalization term}
        \For{$j = i + 1$ \textbf{to} $k-1$}
            \State $\pscore({\scriptstyle \color{blue} T^n[y_{i:k}] \to T^l[y_{i:j}] T^r[y_{j:k}]}) = (\beta_{i:j}[T^l] \cdot \beta_{j:k}[T^r]) / Z  $
        \EndFor
        \texttt{\\}
        \State $\mathit{H} = 0$
        \For{$j = i + 1$ \textbf{to} $k-1$}
            \State $\mathit{H_l} = $ \Call{Recur-Compute-Leaf-Entropy}{$i, j, l, \beta, \dtree$}
            \State $\mathit{H_r}  = $ \Call{Recur-Compute-Leaf-Entropy}{$j, k, r, \beta, \dtree$}
            \LineComment{Denote rule ${\scriptstyle \color{blue} T^n[y_{i:k}] \to T^l[y_{i:j}] T^r[y_{j:k}]}$ as $ {\color{blue} R}$} 
            \State $ \mathit{H} \mathrel{+}= (\mathit{H_l} + \mathit{H_r} - \log [ \pscore({\color{blue} R})]) \cdot \pscore({\color{blue} R}) $
        \EndFor
        \State \Return $\mathit{H}$  \Comment{return the entropy}
    \EndFunction
    \end{algorithmic} 
    \label{algo:leaf-entropy}
\end{algorithm*}

%% file: algo/kl.tex
\begin{algorithm*}[ht]
    \caption{Computing the KL-divergence of $\KL [q(\dtree | \src, \tgt, n) \, \Vert \, p(\dtree | \src, n) ]$}
     \hspace*{3mm} \textbf{Input:} $\pscore_q$: normalized rule scores returned by \Call{Inside-Tree}{$\cdot$} for $q(\dtree | \src, \tgt, n)$, $\pscore_p$: rule scores obtained similarly by calling \Call{Inside-Tree}{$\cdot$} for $p(\dtree | \src, n)$, $n$: number of segments, $\src$: source sentence
    \begin{algorithmic}
    \Function{Compute-Tree-KL}{$\pscore_q, \pscore_p$, $n, \src$}
        \State Initialize $\KL_{\dots}[\cdot] = 0$
        \For{$m = 2$ \textbf{to} $n$}  \Comment number of segments
            \For{$w = 1$ \textbf{to} $|x|$} \Comment width of spans
                \For{$i = 0$ \textbf{to} $|x| - w$}  \Comment start point
                    \State $k = i + w $ \Comment end point
                    \For{$j = i + 1$ \textbf{to} $k-1$}
                        \For{$l = 1$ \textbf{to} $m - 1$} \Comment  number of segments in the left part
                    \State $r = m - l$ \Comment  number of segments in the right part
                            \LineComment{Denote ${\scriptstyle \color{blue} S^m[x_{i:k}] \to I^l[x_{i:j}] S^r[x_{j:k}]}$ as ${\color{blue}R_1}$}
                           \State $\KL_{i, k}[S^m] \hspace{-1mm}\mathrel{+}= \left( \KL_{i:j}[I^l] + \KL_{j:k}[S^r] + \log [\pscore_q({\color{blue}R_1})] - \log [\pscore_p({\color{blue}R_1})] \right) \cdot \pscore_q({\color{blue}R_1}) $
                            \LineComment{Denote $ {\scriptstyle \color{blue} S^m[x_{i:k}] \to I^l[x_{i:j}] I^r[x_{j:k}]}$ as ${\color{blue}R_2}$}
                            \State $\KL_{i, k}[S^m] \hspace{-1mm} \mathrel{+}= \left( \KL_{i:j}[I^l] + \KL_{j:k}[I^r]+ \log [\pscore_q({\color{blue}R_2})] - \log [\pscore_p({\color{blue}R_2})] \right) \cdot \pscore_q({\color{blue}R_2}) $
                            \LineComment{Denote $ {\scriptstyle \color{blue} I^m[x_{i:k}] \to S^l[x_{j:k}] I^r[x_{i:j}]}$ as ${\color{blue}R_3}$}
                            \State $\KL_{i, k}[I^m]\hspace{-1mm} \mathrel{+}= \left( \KL_{j:k}[S^l] + \KL_{i:j}[I^r]+ \log [\pscore_q({\color{blue}R_3})] - \log [\pscore_p({\color{blue}R_3})] \right) \cdot \pscore_q({\color{blue}R_3}) $
                            \LineComment{Denote ${\scriptstyle \color{blue} I^m[x_{i:k}] \to S^l[x_{j:k}] S^r[x_{i:j}]}$ as ${\color{blue}R_4}$}
                            \State $\KL_{i, k}[I^m] \hspace{-1mm} \mathrel{+}= \left( \KL_{j:k}[S^l] + \KL_{i:j}[S^r] + \log [\pscore_q({\color{blue}R_4})] - \log [\pscore_p({\color{blue}R_4})] \right) \cdot \pscore_q({\color{blue}R_4}) $
                        \EndFor
                    \EndFor
                \EndFor
            \EndFor
        \EndFor
        \LineComment{Denote ${\scriptstyle \color{blue} T^n \to S^n[\src]}$, $\quad {\scriptstyle \color{blue} T^n \to I^n[\src]}$ as ${\color{blue}R_5}, {\color{blue}R_6}$ respectively}
        \State $\KL[T^n] \mathrel{+}= \left( \KL_{0:|x|}[S^n] + \log[\pscore_q({\color{blue}R_5})] - \log[ \pscore_p({\color{blue}R_5})]  \right) \cdot \pscore_q({\color{blue}R_5} )$
        \State $\KL[T^n] \mathrel{+}= \left( \KL_{0:|x|}[I^n] + \log[\pscore_q({\color{blue}R_6})] - \log [ \pscore_p({\color{blue}R_6})]   \right) \cdot \pscore_q({\color{blue}R_6})  $ 
        \State \Return $\KL[T^n]$
    \EndFunction
    \end{algorithmic} 
    \label{algo:tree-kl}
\end{algorithm*}